\pdfoutput=1

\documentclass[11pt]{article}

\usepackage[final]{acl}

\usepackage{times}
\usepackage{latexsym}
\usepackage{url}
\usepackage{listings}
\usepackage[normalem]{ulem}
\usepackage{multirow}
\usepackage{subfigure}
\usepackage{alltt,xcolor}

\usepackage[T1]{fontenc}

\usepackage[utf8]{inputenc}

\usepackage{microtype}

\usepackage{inconsolata}
\usepackage{amssymb}
\usepackage{amsmath}
\usepackage{booktabs}
\usepackage{tikz}
\usepackage{pgfplots}
\usepackage[english]{babel}

\title{\datasetname: Quantifying Uncertainty in Natural Language Text \\ in Bayesian Reasoning Scenarios}

\author{Timo Pierre Schrader$^{1,2}$~
  Lukas Lange$^1$~
  Simon Razniewski$^{3}$~
  Annemarie Friedrich$^2$\\
  $^1$Bosch Center for Artificial Intelligence, Renningen, Germany \\ 
    $^2$University of Augsburg, Augsburg, Germany \\
    $^3$ScaDS.AI \& TU Dresden, Dresden, Germany \\
\texttt{timo.schrader|lukas.lange@de.bosch.com} \\
\texttt{simon.razniewski@tu-dresden.de} \\
  \texttt{annemarie.friedrich@informatik.uni-augsburg.de}}

\usepackage{soul}
\usepackage{todonotes}
\usepackage{amsmath}
\usepackage{enumitem}
\usepackage{placeins}
\usepackage{csquotes}
\usepackage{tikz}
\usepackage{bbm}

\newcommand{\tref}[1]{Table~\ref{#1}}

\usepackage{soul}
\usepackage{xspace}

\setlist[description]{leftmargin=\parindent,labelindent=0pt,itemsep=0pt}

\definecolor{darkgreen}{rgb}{0.0, 0.5, 0.0}

\newcommand{\red}[1]{\textcolor{red}{#1}}

\newcommand{\green}[1]{\textcolor{darkgreen}{#1}}
\newcommand{\purple}[1]{\textcolor{purple}{#1}}

\newcommand{\datasetname}[0]{\textsc{Quite}\xspace}
\newcommand{\blind}[0]{BLInD\xspace}
\newcommand{\cladder}[0]{\textsc{Cladder}\xspace}

\newcommand{\causalcot}[0]{\textsc{CausalCoT}\xspace}

\usepackage{amssymb}
\usepackage{utfsym}
\newcommand{\crossmark}{\scalebox{0.75}{\usym{2613}}}

\newenvironment{customlegend}[1][]{%
\begingroup
\csname pgfplots@init@cleared@structures\endcsname
\pgfplotsset{#1}%
}{%
\csname pgfplots@createlegend\endcsname
\endgroup
}%
\def\addlegendimage{\csname pgfplots@addlegendimage\endcsname}
\pgfplotsset{
cycle list={%
{mark=*,solid,semithick,line legend, color=blue},
{mark=square*,solid, semithick, color=red},%
{mark=pentagon*,solid,semithick,color=violet},%
{mark=triangle*, solid, semithick,color=orange},
{mark=diamond*,solid,semithick, color=cyan},
{mark=triangle,solid, color=black},
{mark=star,solid, color=violet},
{mark=pentagon,solid, color=teal}
}}

\begin{document}
\maketitle
\begin{abstract}
Reasoning is key to many decision making processes.
It requires consolidating a set of rule-like premises that are often associated with degrees of uncertainty and observations to draw conclusions.
In this work, we address both the case where premises are specified as numeric probabilistic rules and situations in which humans state their estimates using words expressing degrees of certainty.
Existing probabilistic reasoning datasets simplify the task, e.g., by requiring the model to only rank textual alternatives, by including only binary random variables, or by making use of a limited set of templates that result in less varied text.

In this work, we present \datasetname, a question answering dataset of real-world Bayesian reasoning scenarios with categorical random variables and complex relationships.
\datasetname provides high-quality  natural language verbalizations of premises together with evidence statements, and expects the answer to a question in the form of an estimated probability.
We conduct an extensive set of experiments, finding that logic-based models outperform out-of-the-box large language models on all reasoning types (causal, evidential, and explaining-away).
Our results provide evidence that neuro-symbolic models are a promising direction for improving complex reasoning. %
We release \datasetname and code for training and experiments on Github.\footnote{\url{https://github.com/boschresearch/quite-emnlp24}}

\end{abstract}

\section{Introduction}

Reasoning about causality is an integral part of intelligence, as it helps to understand and predict the world.
In the real world, causes and associations can rarely be determined with complete certainty, and reasoning becomes inherently difficult if uncertainties are involved \citep{DBLP:books/daglib/0066829}.
An automated system for interpreting text describing causal relationships and their associated numeric probabilities or verbalized degrees of uncertainty would be highly useful in domains such as requirements engineering \citep{DBLP:conf/re/YangRGWN12} or text-mining in clinical documentation \cite{DBLP:conf/clef/TurnerIV21}. %
Modeling linguistically expressed uncertainty has been an active research area for decades in natural language processing (NLP) \citep{szarvas-etal-2008-bioscope,DBLP:conf/wims/JeanHRBM16,sileo2023probing}.

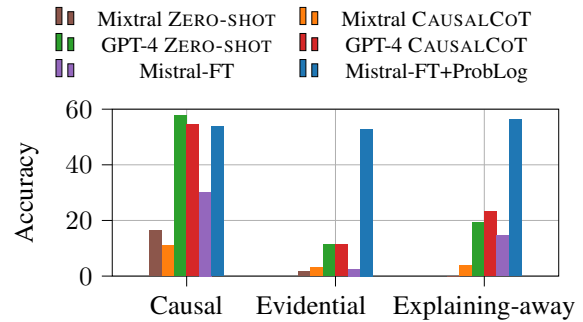
\begin{figure}[t]
    \centering
    \resizebox{0.85\linewidth}{!}{\begin{tikzpicture}

\definecolor{crimson2143940}{RGB}{214,39,40}
\definecolor{darkgray176}{RGB}{176,176,176}
\definecolor{darkorange25512714}{RGB}{255,127,14}
\definecolor{forestgreen4416044}{RGB}{44,160,44}
\definecolor{lightgray204}{RGB}{204,204,204}
\definecolor{mediumpurple148103189}{RGB}{148,103,189}
\definecolor{sienna1408675}{RGB}{140,86,75}
\definecolor{steelblue31119180}{RGB}{31,119,180}

\begin{customlegend}[legend columns=2,legend style={align=center,draw=none,column sep=2ex},
width=4.8in,
legend entries={Mixtral \textsc{Zero-shot}, Mixtral \textsc{CausalCoT}, GPT-4 \textsc{Zero-shot}, GPT-4 \textsc{CausalCoT}, Mistral-FT, Mistral-FT+ProbLog}
]
\addlegendimage{ybar,ybar legend,draw=none,fill=sienna1408675}
\addlegendimage{ybar,ybar legend,draw=none,fill=darkorange25512714}
\addlegendimage{ybar,ybar legend,draw=none,fill=forestgreen4416044}
\addlegendimage{ybar,ybar legend,draw=none,fill=crimson2143940}
\addlegendimage{ybar,ybar legend,draw=none,fill=mediumpurple148103189}
\addlegendimage{ybar,ybar legend,draw=none,fill=steelblue31119180}
\end{customlegend}
\end{tikzpicture}}
    \resizebox{1.0\linewidth}{!}{\begin{tikzpicture}

\definecolor{crimson2143940}{RGB}{214,39,40}
\definecolor{darkgray176}{RGB}{176,176,176}
\definecolor{darkorange25512714}{RGB}{255,127,14}
\definecolor{forestgreen4416044}{RGB}{44,160,44}
\definecolor{lightgray204}{RGB}{204,204,204}
\definecolor{mediumpurple148103189}{RGB}{148,103,189}
\definecolor{steelblue31119180}{RGB}{140,86,75}
\definecolor{sienna1408675}{RGB}{31,119,180}

\begin{axis}[
legend cell align={left},
legend style={fill opacity=0.8, draw opacity=1, text opacity=1, draw=lightgray204},
tick align=outside,
tick pos=left,
x grid style={darkgray176},
xmin=-0.5, xmax=2.5,
xtick style={color=black},
xtick={0,1,2},
xticklabel style={rotate=0.0},
xticklabels={Causal,\hspace{-0.7cm}Evidential,Explaining-away},
y grid style={darkgray176},
ymin=0, ymax=60.0,
ytick style={color=black},
ylabel={Accuracy},
grid,
legend pos=outer north east,
height=4cm,
width=8cm
]
\draw[draw=none,fill=steelblue31119180] (axis cs:-0.25,0) rectangle (axis cs:-0.166666666666667,16.3);

\draw[draw=none,fill=steelblue31119180] (axis cs:0.75,0) rectangle (axis cs:0.833333333333333,1.61);
\draw[draw=none,fill=steelblue31119180] (axis cs:1.75,0) rectangle (axis cs:1.83333333333333,0);
\draw[draw=none,fill=darkorange25512714] (axis cs:-0.166666666666667,0) rectangle (axis cs:-0.0833333333333333,10.87);

\draw[draw=none,fill=darkorange25512714] (axis cs:0.833333333333333,0) rectangle (axis cs:0.916666666666667,3.23);
\draw[draw=none,fill=darkorange25512714] (axis cs:1.83333333333333,0) rectangle (axis cs:1.91666666666667,3.85);
\draw[draw=none,fill=forestgreen4416044] (axis cs:-0.0833333333333333,0) rectangle (axis cs:-1.38777878078145e-17,57.61);

\draw[draw=none,fill=forestgreen4416044] (axis cs:0.916666666666667,0) rectangle (axis cs:1,11.29);
\draw[draw=none,fill=forestgreen4416044] (axis cs:1.91666666666667,0) rectangle (axis cs:2,19.23);
\draw[draw=none,fill=crimson2143940] (axis cs:-3.46944695195361e-17,0) rectangle (axis cs:0.0833333333333333,54.35);

\draw[draw=none,fill=crimson2143940] (axis cs:1,0) rectangle (axis cs:1.08333333333333,11.29);
\draw[draw=none,fill=crimson2143940] (axis cs:2,0) rectangle (axis cs:2.08333333333333,23.08);
\draw[draw=none,fill=mediumpurple148103189] (axis cs:0.0833333333333333,0) rectangle (axis cs:0.166666666666667,30.21);

\draw[draw=none,fill=mediumpurple148103189] (axis cs:1.08333333333333,0) rectangle (axis cs:1.16666666666667,2.58);
\draw[draw=none,fill=mediumpurple148103189] (axis cs:2.08333333333333,0) rectangle (axis cs:2.16666666666667,14.62);
\draw[draw=none,fill=sienna1408675] (axis cs:0.166666666666667,0) rectangle (axis cs:0.25,53.7);

\draw[draw=none,fill=sienna1408675] (axis cs:1.16666666666667,0) rectangle (axis cs:1.25,52.9);
\draw[draw=none,fill=sienna1408675] (axis cs:2.16666666666667,0) rectangle (axis cs:2.25,56.15);
\end{axis}
\end{tikzpicture}}
    \vspace{-0.2cm}
    \caption{Percentage of instances solved correctly for each \textbf{Bayesian reasoning type}. 
    The neuro-symbolic Mistral-FT+ProbLog approach  is robust against the inherent difficulties of  different reasoning types.}
    \label{fig:reasoning_types}
    \vspace{-0.3cm}
\end{figure}

\begin{figure*}[!t]
    \centering
    \vspace{-0.3cm}
    \includegraphics[width=1.0\linewidth]{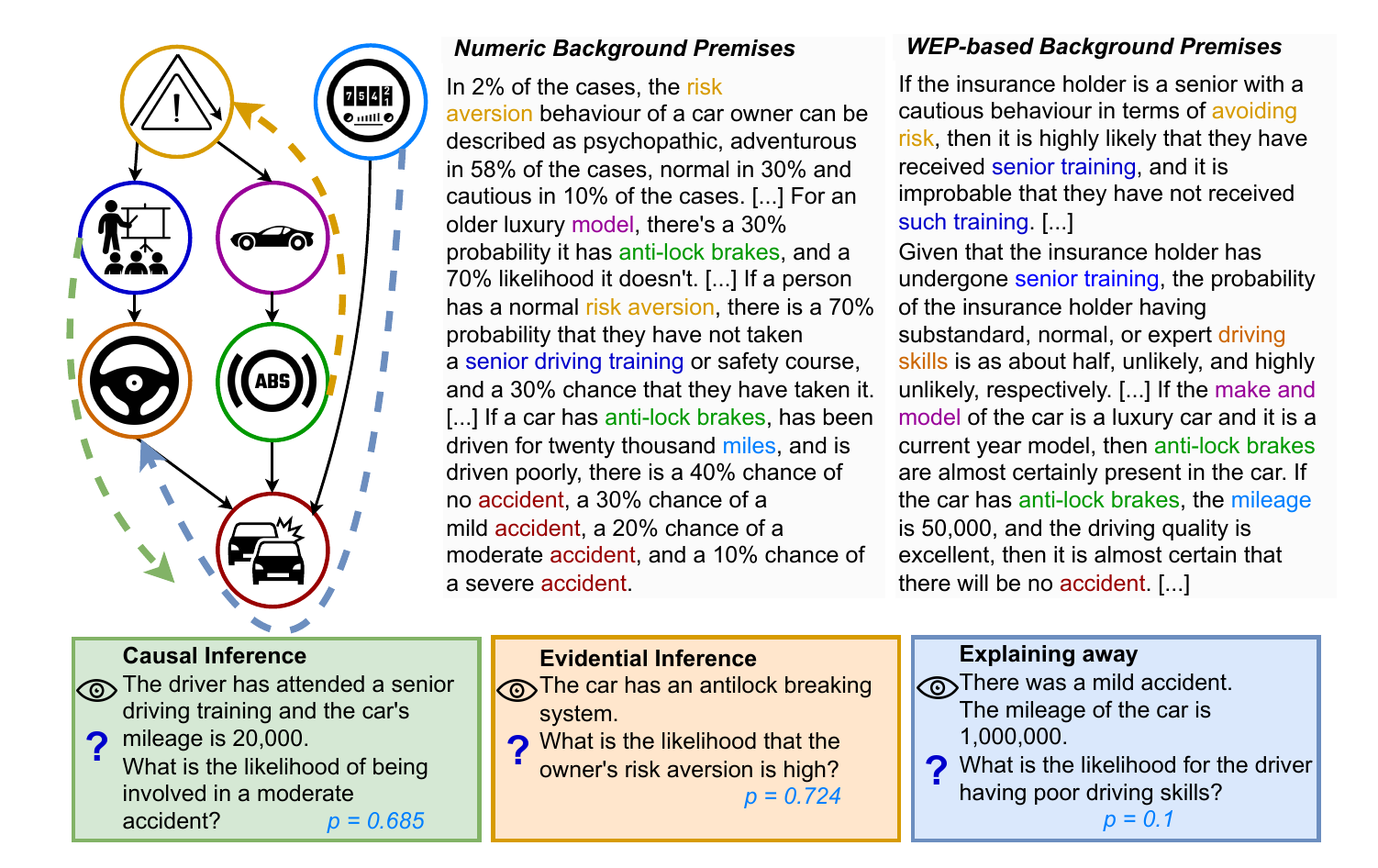}
    \begin{verbatim}
    \end{verbatim}
    \vspace{-1cm}
    \caption{Example instances of \datasetname. Each question is categorized according to the reasoning pattern. %
    }
    \label{fig:teaser}
\end{figure*}

Recently, large language models (LLMs) have shown superior performance on many NLP tasks.
However, they fall short of incorporating principled reasoning mechanisms, with frequent failure cases \citep{DBLP:journals/corr/abs-2305-00050}, and their mathematical skills decline if presented with unseen cases \citep{frieder2023mathematical,yousefzadeh2023large}.
In zero-shot or chain-of-thought (CoT) prompting settings, open-source and open-weights LLMs are also unable to outperform a random baseline in Bayesian inference in higher-order causal networks \citep{jin2023cladder}.
GPT-3 and GPT-4 are somewhat better, yet do not excel at the task.

To probe LLMs for their reasoning capabilities, \citet{jin2023cladder} compile the \cladder~dataset based on toy causal inference scenarios taken from textbooks and literature on causal reasoning.
\cladder~evaluates performance by \textit{rungs} of the \textit{ladder of causation} \citep{bookofwhy}.\footnote{The rungs of the ladder are: (1) statistical dependencies based on observations: \enquote{If I am vaccinated, how likely am I to survive?} (2) interventions: \enquote{If I get vaccinated, what is the likelihood of surviving?} (3) counterfactual reasoning: \enquote{Would a person have survived if they had been vaccinated?}}
While this work inspired ours, it suffers from several limitations: First, its networks only include binary random variables. Second, questions are of the form \enquote{Does X increase the likelihood of Y?} and expect yes/no answers (with a 50:50 distribution). Hence, it is not designed to estimate the correlation of model output with the probabilities estimated according to the Bayesian network.

In this paper, we present \datasetname, a new benchmark for \textbf{Q}uantifying \textbf{U}ncertainty \textbf{i}n natural language \textbf{Te}xt.
As illustrated in Figure \ref{fig:teaser}, \datasetname goes one step further, leveraging toy and real-world causal networks and asking the model to output a much finer-grained numeric probability estimate.
In addition, our work is the first to make use of categorical random variables (and not just binary variables as in existing related datasets).
Despite using real-world networks, our dataset is not solvable from a question-evidence baseline alone, which demonstrates that the model cannot solve the task solely from background knowledge acquired during pre-training.
To the best of our knowledge, our work is the first to explicitly distinguish the \textbf{three Bayesian inference types} \textit{causal inference}, \textit{evidential reasoning}, and \textit{explaining-away}, which directly reflect the reasoning paths in the network.\footnote{We address them in situations corresponding to rung 1 of the causal ladder.}

Following the recent trend of probing LLMs for their mathematical capabilities, the \blind~ dataset \citep{nafar2024probabilistic} focuses on the numeric Bayesian reasoning capabilities of GPT models, hence only uses dummy event variables (e.g., \enquote{orange event}).
\citeauthor{nafar2024probabilistic} find that using program-aided language models \citep{DBLP:conf/icml/GaoMZ00YCN23} and neuro-symbolic approaches can drastically increase model performance.
In their work, however, it remains an open research question whether this approach scales to less template-like and more varied natural language text.
\datasetname goes one important step into this direction: with the support of LLMs, we verbalize complex real-world Bayesian reasoning scenarios in a linguistically more varied style.
Our validation shows that instances in \datasetname are of high accuracy with regard to probabilistic information, and use more complex yet mostly error-free language compared to existing datasets.

Existing datasets focus on probing for Bayesian reasoning capabilities when presented with verbalized \textit{numeric} conditional probability tables.
\datasetname also offers a setting that mimicks human conversation, replacing probabilities with \textit{words of estimative probability} (WEPs) such as \enquote{unlikely} or \enquote{improbable.}
This scenario has previously only been investigated in the context of natural language inference \citep{sileo2023probing}.
While in this setting, parsing is more difficult for all models, performance is encouraging.
Our results illustrate that when targeting natural text, structured causal models in combination with LLMs are a promising approach to estimating likelihood.

A highly interesting finding of our experimental study is that all included LLMs (including GPT models) fail on questions requiring evidential and explaining-away reasoning both in zero-shot and CoT settings (see Figure \ref{fig:reasoning_types}).
We hypothesize that the LLMs in our study have learned a good concept of causality during pre-training, and that causal reasoning scenarios can often be solved based on statistical patterns.
Potentially, LLMs incorporate biases for assuming causal analyses, as these types of relationships are more frequently expressed in pretraining data.
By contrast, our experimental results demonstrate that a fine-tuned neuro-symbolic system has no difficulties solving the latter two categories as well.
We hence conclude that for integrating complex reasoning capabilities into NLP systems, neuro-symbolic models are a promising (if not necessary) direction.

Our contributions are as follows:
(1) We present a novel dataset of verbalizations of Bayesian networks including categorical variables in two versions (with explicit probabilities vs.~words of estimative probability), symbolic target representations, and question-evidence pairs that provide simulated observations and queries asking for probabilities.
(2) To the best of our knowledge, all closely related recent prior work uses either vanilla or CoT LLMs, or generates neuro-symbolic representations simply via manually designed prompts. Our work is the first to explicitly fine-tune state-of-the-art LLMs on semantic parsing to probabilistic first-order programming language ProbLog \citep{problog,problog2}. It consistently and strongly outperforms purely LLM-based approaches in probabilistic reasoning on \datasetname.

\section{Background and Related Work}
In this section, we start by introducing the basic concepts of Bayesian networks and corresponding reasoning types, closely following the terms and definitions of \citet{koller2009probabilistic}.
We then review existing literature on modeling uncertainty in language, benchmark datasets for Bayesian reasoning, and semantic parsing to logical forms.

\paragraph{Bayesian Networks and Reasoning Patterns.}%
Bayesian networks (BNs) represent joint probability distributions over a set of random variables and probabilistic dependencies between them.
These networks are modelled as directed acyclic graphs with nodes and directed edges between the nodes.
Nodes represent random variables that can take two or more states.
Edges correspond to probabilistic dependencies that are represented in so-called \textit{conditional probability tables} (CPTs).
A random variable $X_i$ is an observable attribute that can randomly take two or more disjoint states.
For example, the outcome of throwing a coin could be either \textit{head} or \textit{tail}.
A Bayesian network therefore represents a joint probability distribution over a set of random variables $\{X_1, \dots, X_n\}$: $\mathbb{P}(X_1,\dots,X_n)$.
A conditional probability distribution, denoted by edges in the network, is a modification of the joint probability in which a random variable is conditioned on one or more other random variables: $\mathbb{P}(X_i|X_j,\dots)$.
Hence, there is now a dependency between $X_i$ and all its parents in the graph, i.e., the value of the parents directly influences the outcome of $X_i$.

The combined nature of directed edges allows for different reasoning patterns.
\textit{Causal} reasoning requires drawing conclusions about an effect if its cause is observed.
Vice versa, reasoning about the cause of an observed effect is called \textit{evidential} reasoning.
Finally, drawing conclusions about a cause if an effect and further causes of this effect are observed is called \textit{explaining-away}.
For a more in-depth introduction to Bayesian networks and reasoning patterns, please refer to \autoref{sec:theoretical_background}.

\paragraph{Modeling uncertainty.}
BioScope \citep{szarvas-etal-2008-bioscope,farkas-etal-2010-conll,vincze-2010-speculation} is an early work addressing the modeling of uncertainty in biomedical text by marking triggers and their scope. A cluster of works has focused on modal verbs which are a frequent trigger \citep{ruppenhofer-rehbein-2012-yes,zhou-etal-2015-semantically,henning-etal-2022-mist,wagner-zarriess-2023-probing,owan-etal-2023-quirk}.

Recently, much research concentrates on probing how LLMs react to prompts containing expressions of (un)certainty.
\citet{zhou-etal-2023-navigating} find that LLMs are highly sensitive to epistemic markers of certainty in the prompt, decreasing question answering (QA) performance drastically.
Conversely, models have been tested with regard to whether they can express their own confidence in an answer \citep{tian-etal-2023-just}.

Modeling uncertainty has been investigated using Natural Language Inference (NLI) tasks.
\citet{sileo2023probing} frame uncertainty-based reasoning as NLI to study how LLMs deal with \textit{words of estimative probability} (WEP) such as \enquote{likely} or \enquote{improbable.}
The task of Uncertain NLI (UNLI) \citep{chen-etal-2020-uncertain} targets predicting a numeric score for the uncertainty in entailment between two (non-quantified) statements.
\citet{talman-etal-2023-uncertainty} explicitly model the variation in judgments of NLI instances exhibited by groups of annotators.

\paragraph{Bayesian reasoning: benchmark datasets.}
\cladder~\cite{jin2023cladder} consists of 10k instances of verbalized Bayesian networks and associated questions.
Their \textit{stories} provide an overall summary of the direct effects and spell out the CPTs.
The BLInD dataset \citet{nafar2024probabilistic} tests to what extent GPT-3.5 and GPT-4 can perform Bayesian reasoning using template-based descriptions of dummy events.
In contrast, \datasetname focuses on complex real-world scenarios.
\cladder~ and \blind~instances are verbalized exclusively based on templates, while \datasetname exhibits a much larger linguistic variety and higher grammaticality. 
Key differences between the three datasets are summarized in Table \ref{tab:dataset_comparison}.
All three studies (including ours) find that basic QA prompting does not work very well, but that COT prompting brings improvements.
One example instance from of each dataset is provided in \autoref{sec:related_datasets}.

\begin{table}[t]
    \centering
    \footnotesize
    \setlength{\tabcolsep}{4pt}
    \begin{tabular}{l|ccc}
    \toprule
     & \textbf{\datasetname} & \textbf{\cladder} & \textbf{\blind}\\
    \midrule
    \textit{Categorical variables} & \green{\checkmark} & \red{\crossmark} & \red{\crossmark}\\
    \textit{Rung of causations} & \red{\crossmark} & \green{\checkmark} & \red{\crossmark}\\
    \textit{WEP-based uncertainty} & \green{\checkmark} & \red{\crossmark} & \red{\crossmark}\\
    \textit{ProbLog representations} & \green{\checkmark}  & \red{\crossmark} & \green{\checkmark} \\
    \textit{Topic/Domain variety} & \green{\checkmark} & \green{\checkmark}  & \red{\crossmark} \\
    \bottomrule
    \end{tabular}
    \caption{Comparison of \datasetname, \cladder, and \blind.}
    \label{tab:dataset_comparison}
\end{table}

\paragraph{Semantic parsing to logical form.}
Constructing structured representations from natural language text has been a long-standing research area in NLP \citep{zettlemoyer-collins-2007-online,reddy-etal-2016-transforming,kim-etal-2021-transition}.
Recent work involves the specialization of LLMs on this task.
\citet{linc} present a framework called \textit{LINC} that translates logical statements into domain-specific languages, where the LLM acts as semantic parser and bridges the gap between natural language and structured, neuro-symbolic representations.
\citet{satlm} %
employ an LLM to generate declarative sets of rules that are handed over to a SAT solver executable.
\citet{nafar2024probabilistic} prompt LLMs to generate symbolic ProbLog code for solving %
the \blind dataset. %

\section{Dataset}
In this section, we describe the dataset creation process of \datasetname, 
provide dataset statistics, assess its linguistic quality, and validate the translations of the logical structures into natural language. %

\subsection{Dataset Structure}
QA instances in \datasetname are composed from the following parts that facilitate testing model behavior when performing Bayesian inference and reasoning with uncertainty.
For an example, see Figure \ref{fig:teaser}.%

\begin{description}
    \item[Numeric background premises] %
    are verbalizations of CPTs explicitly mentioning percentages.: \textit{In 2\% of the cases, the risk aversion behaviour of a car owner can be described as psychopathic [...]}
    \item[WEP-based background premises] are verbalizations of CPTs replacing every numeric probability value by an uncertainty quantifier (cf. Section \ref{ssec:non-numeric-premises}): \textit{There is almost no chance that the risk aversion behaviour of a car owner can be described as psychopathic [...]}
    \item[Question-evidence (QE) pairs:] Evidences are observations that set the value of one or multiple random variables in the Bayesian model to a particular value. Queries are then asking for the probability of a single random variable that is inferable given the evidence, i.e., $\mathbb{P}(X_Q = x_q^{m_q} | X_{E_1} = x_{E_1}^{m_1}, \dots, X_{E_j} = x_{E_j}^{m_j})$, where $x_{E_i}^{m_i}$ refers to the value that is assigned to random variable $X_{E_i}$.
\end{description}

\subsection{Data Collection}
Our dataset is composed from a collection of publicly available BNs compiled from the literature.
They reflect realistic probabilistic relationships in several domains (medicine, severe weather forecasting, car insurance, mildew growth, phytophthora species, protein signalling, water treatment, and software troubleshooting).
Our first data source is the \textit{bnlearn} library \citep{bnlearn}, %
which is commonly used in benchmarking scenarios for algorithms for BNs \citep{10.5555/3586589.3586913, 10.5555/1641503.1641512, DBLP:journals/corr/abs-1210-5135}.
Our second source is the \textit{BNMA BN repository}.\footnote{\url{https://www.abnms.org/bnrepo/}}
Some of the BNs contain node counts in the order of magnitude $100$ or $1000$, hence, to keep the networks manageable in terms of size, we split them into subnetworks.
We marginalize %
root nodes in subnetworks if they have ancestors in the larger original network to obtain self-contained BNs. %
An example for this process is provided in \autoref{sec:bn_subsetting}.
We end up with a total of 14 different BNs, split into a total of 30 subnetworks.
Our BNs contain nodes of degree $0$ to $3$, i.e., there are zero to three conditions (parent nodes) on which a random variable can depend. %
To the best of our knowledge, we are the first to verbalize these widely known networks to natural language, thereby making them available as a resource for NLP research.

\subsection{Dataset Creation Steps}
We semi-automatically create the natural language part of the dataset with the help of LLMs as illustrated in \autoref{sec:data_generation_pipeline}.
As LLM backbone in our pipeline, we use Mixtral-8x7B-Instruct-v0.1 \citep{jiang2024mixtral}.\footnote{We also experiment with Mixtral-8x22B-Instruct-v0.1, but due to budget reasons, we mostly stick to Mixtral-8x7B-Instruct-v0.1. A small percentage (ca. $8\%$) of the dataset has been created manually.}
Each background premise in the dataset describes the probabilities (either expressed numerically or using WEPs) for all possible assignments to one random variable $X_i$, given one specific assignment to all the conditions.
We generate template-based premises by iterating over every entry of each CPT and fill templates of the form \textit{If [Conditions], then [Probabilities]}, where conditions refer to all incoming edges in the Bayesian network and probabilities to the currently selected variable.
Next, to create natural language premises, we prompt Mixtral with a prompt containing technical explanations of the variables in the network, few-shot examples as well as the template-based premises.
In contrast to related datasets that fully rely on rule-based templates, \datasetname hence contains more varied descriptions.%
For each network, we create a representation in ProbLog in a semi-automatic way: We manually define predicates for all nodes and categories in the CPTs, and then use a rule-based conversion.
For the entire ProbLog data (1192 statements), the first author of the paper has manually checked if the statements match their ProbLog counterparts and if the wording and use of domain-specific vocabulary is consistent throughout the entire verbalized network.
This was an extensive manual effort of multiple months.
The ProbLog representation enables us to perform fine-tuning of the semantic parser.

\subsubsection{WEP-based Background Premises}
\label{ssec:non-numeric-premises}
To guide the LLM to express natural language uncertainty in a principled way, we rely on a human study conducted by \citet{fagen} that includes the subjective judgements by more than 100 people who were asked to judge which numeric probability they associate with each adverb in a list.
These adverbs are often referred to as \textit{words of estimative probability} (WEP), %
a term that mainly originates from the work of \citet{kent1964words}, which investigates the mapping between specific uncertainty quantifiers and probabilities (see \autoref{sec:wep}).
We map  each probability value to the closest adverb.
In case there is more than one possible adverb (e.g., 10\% maps to \textit{improbable}, \textit{little chance}, and \textit{chances are slight}), one of them is randomly selected.
Additionally, to simulate subjectivity, we %
select the second-closest adverb in 10\% of the cases.
If all states of a random variable have the same probability, we manually correct the verbalization to \enquote{equally likely.}

The heuristic of choosing the WEPs based on the premises' probabilities works well in most cases, yet we observe that this heuristic does not fully fit cases where all states of a categorical random variable have a low probability.
For example, assume we have $\mathbb{P}(X_i = x_i^1) = 0.2$, $\mathbb{P}(X_i = x_i^2) = 0.2$, $\mathbb{P}(X_i = x_i^3) = 0.3$, and $\mathbb{P}(X_i = x_i^4) = 0.3$.
This would lead to the following verbalization: \textit{It is probably not the case that $X_i$ takes the value $x_i^3$ or $x_i^4$, and it is unlikely that it takes $x_i^1$ or $x_i^2$.}
We manually add additional information to these instance describing the state that is still the most likely one.
We leave the adaption of WEPs to this edge case as a direction for future research.

\subsubsection{Question-Evidence (QE) Pairs}
We construct QE pairs as follows. %
As evidences, we randomly sample 1 to n-1 observations per instance (i.e., $X_{E_1} = x_{E_1}^{m_1}, \dots, X_{E_j} = x_{E_j}^{m_j}$) and let Mixtral transform them to natural language statements such as \enquote{The accident was mild.}
For the question, we sample one node $X_q$ for which Mixtral formulates a question of the form: \enquote{What is the likelihood of $X_q$ having the value $x_Q^m$?}
Each QE pair requires calculating the probability $\mathbb{P}(X_Q = x_Q^m | X_{E_1} = x_{E_1}^{m_1}, \dots, X_{E_j} = x_{E_j}^{m_j})$. %
The ground truth answer (numeric probability value) is calculated based on the underlying probabilistic model of each subnetwork.
Most QE instances can be clearly categorized into their respective reasoning pattern.
We determine \textit{causal} and \textit{evidential} reasoning by inspecting the list of parent and child nodes, respectively, and check if one of them is observed, i.e., part of the evidence.
To identify explaining-away QEs, we only check if one of the direct child nodes of $X_Q$ and one of their direct parents is observed.
The test set contains 92 causal, 62 evidential and 26 explaining-away QE instances.

\subsection{Dataset Statistics}
We provide detailed statistics for \datasetname in \tref{tab:statistics}.
We ensure that subnetworks that are derived from the same original network are assigned to only training or test data, respectively.
All QE pairs have been manually checked and if necessary corrected by the first author.
On average, there are three to four states per random variable. %
The average number of background premises reflects the amounts of probabilistic statements that need to be processed before reasoning.
The average number of premises per network in \datasetname is much higher than those in related works, reflecting its challenging nature.
The statistics in the lower part of \tref{tab:statistics} differ between training and test split due to taking the original networks into account.

\begin{table}[t]
    \centering
    \footnotesize
    \setlength{\tabcolsep}{3pt}
    \begin{tabular}{l|rrr}
    \toprule
     & \textbf{\# Train} & \textbf{\# Test} & \textbf{Total} \\
    \midrule
    \textit{Networks} & 20 & 10 & 30 \\
    \textit{Numeric premises} & 930 & 273 & 1192\\
    \textit{WEP-based premises} & 930 & 273 & 1192\\
    \textit{Evidence statements} & 812 & 578 & 1390\\
    \textit{Queries} & 347 & 230 & 577\\
    \midrule
    \textit{Avg. \# states / rand. var.} & 3.5$_{\pm 2.0}$ & 2.9$_{\pm 1.4}$ & 3.3$_{\pm 1.9}$ \\
    \textit{Avg. \# rand. var. / net} & 5.9$_{\pm 2.4}$ & 6.0$_{\pm 2.5}$ & 5.9$_{\pm 2.4}$ \\
    \textit{Avg. \# premises / net} & 46.5$_{\pm 31.2}$  & 27.3$_{\pm 23.2}$ & 40.1$_{\pm 30.2}$ \\
    
    \bottomrule
    \end{tabular}
    \caption{Dataset statistics for \datasetname. Subscripts denote standard deviation.}
    \label{tab:statistics}
    \vspace{-0.5cm}
\end{table}

\subsection{Validation and Quality Assessment}

In contrast to QE pairs, premise statements are assembled in a semi-automatic fashion.
In this section, we validate their correctness and examine the linguistic quality of the entire dataset.

\paragraph{Validation.}
\label{sssec:validation}
The first author of the paper has performed extensive checking and correcting for the 2384 premise statements.
As a second validation step, two of the (non-first) authors of this paper that were not exposed to the generation process before are presented with 400 randomly sampled premises of \datasetname (200 numeric and 200 WEP-based premises).
They are asked to assess whether the LLM-generated output contains all input variables and states and whether probabilities have been translated correctly.
Of the numeric premises, 193 instances (96.5\%) correctly describe the underlying probability distribution without ambiguities.
Most errors relate to rounding close-to-zero probabilities to zero.
Of the WEP-based instances, 188 instances (94\%) are correct, with the LLM misinterpreting the input and wrong representations of the probability values being the main error causes.
Our validation study shows that \datasetname contains mostly well-formed instances that correctly reflect the random variable states and probabilities.

\paragraph{Linguistic Quality Assessment.}
To assess the linguistic quality of \datasetname, we make use of Grammarly,\footnote{\url{https://app.grammarly.com/}} a state-of-the-art commercial writing assistant.
We compare \datasetname to \cladder(excluding its non-sensical subset) %
and \blind.
We randomly sample premises, evidences, and queries until a character count of approximately 95,000 has been reached. %

Results are provided in \tref{tab:grammarly}.
On average, \datasetname has much fewer grammar and spelling mistakes per instance than \cladder.\footnote{\blind templates lack determiners: \enquote{If purple event is False, then grey event is True with probability of 39\%.}%
} %
This highlights the advantage of LLM-based generation over template-based instance generation.
\datasetname has the most specific vocabulary, demonstrated by the highest amount of rare words, which Grammarly defines as words that do not belong to the 5k most frequent English words. %
According to the Flesh-Kincaid readability score \citep{kincaid1975derivation}, \blind requires skills of the level of 8th/9th graders, \cladder and the WEP-based part of \datasetname need 10th to 12th grade skills, and the numeric part of \datasetname requires college-level reading skills. %

Finally, \datasetname edges out on the two other datasets in terms of the overall Grammarly score.
This as a strong indicator that our dataset is much closer to human-like natural language.
We conclude from this analysis that our dataset makes use of rich language with a complex vocabulary, and is close to human-like language.
Overall, \datasetname is well-suited for assessing the reasoning capabilities of state-of-the-art models in realistic scenarios.

\begin{table}[t]
    \centering
    \footnotesize
    \setlength{\tabcolsep}{3pt}
    \begin{tabular}{l|cccc}
    \toprule
    \multirow{2}[2]{*}{\textbf{Grammarly Metric}} & \multicolumn{2}{c}{\textbf{\datasetname}} & \multirow{2}[2]{*}{\textbf{\textsc{Clad.}}} & \multirow{2}[2]{*}{\textbf{\blind}} \\
    \cmidrule(lr){2-3}
    & Num. & WEP & & \\
    \midrule
    \textit{Writing Issues / Instance} & 0.3 & 0.3 & 1.3 & 30.9 \\
    \textit{Rare words} & 43\% & 41\% & 36\% & 23\%\\
    \textit{Readability} & 48 & 50 & 50 & 68\\
    \midrule
    \midrule
    \textbf{Overall judgement} (0-100) & \textbf{85} & \textbf{82} & 45 & 34 \\
    \bottomrule
    \end{tabular}
    \caption{Dataset quality assessment by Grammarly for a randomly sampled subset of all datasets. }
    \label{tab:grammarly}
    \vspace{-0.2cm}
\end{table}

\section{Modeling }
\label{sec:modelling}
In this section, we describe the various models that we benchmark using \datasetname.
Fine-tuned models are suffixed \textit{-FT} in the following.

\subsection{LLM Prompting Methods}
We experiment with several prompting techniques for state-of-the-art LLMs of different sizes.\footnote{GPT4-Turbo (\textit{turbo-2024-04-09}) \citep{openai2024gpt4}, Llama-3-8B-Instruct \citep{llama3modelcard}, Mistral-7B-Instruct-v0.3 \citep{jiang2023mistral}, and Mixtral-8x7B-Instruct-v0.1 \citep{jiang2024mixtral}. The temperature is set to $0.0$.}
In the \textbf{zero-shot} setting, we provide all background premises of the network, a set of evidences and a question asking for the probability of a specific random variable taking a selected state.\footnote{We also performed preliminary experiments with an one-shot example %
which did not result in consistent improvements.}
The \textbf{\causalcot} technique was introduced by \citet{jin2023cladder} and asks the model to build up the probabilistic graph, to extract the question type and to perform the mathematical calculation step by step.

\subsection{Neuro-symbolic Approach}
Our ProbLog-based approach separates problem understanding and probabilistic reasoning, first parsing each premise (both numeric and WEP-based), evidence statements, and queries into a ProbLog program.
In logic programming languages, declarative programs are defined as a series of rules and facts that, in combination, evaluate to true or false. 
In Prolog, rules are defined in first-order logic, where a rule body defines which conditions need to be met (i.e., need to be \textit{true}) in order for the rule head to be evaluated as true.
ProbLog \citep{problog,problog2}, which we use in this work, is a \textbf{probabilistic programming language} that extends the functionality of Prolog.
It allows the specification of probabilistic models by declaring the probability distributions in FOL-style formulas.
Since our dataset comprises not only binary, but also categorical random variables, we use annotated disjunctions, e.g.,

\begin{small}
\begin{alltt}
    0.02::\purple{risk_aversion}(car_owner, psychopathic);
    0.58::\purple{risk_aversion}(car_owner, adventurous);
    0.30::\purple{risk_aversion}(car_owner, normal);
    0.10::\purple{risk_aversion}(car_owner, cautious).
\end{alltt}
\end{small}
    
We first parse background premises into ProbLog using either a zero-shot LLM (\textbf{ProbLog-Prompt}) or an LLM fine-tuned for text-to-ProbLog parsing (\textbf{ProbLog-FT}). %
The QE pairs are parsed as a second step (i.e., after the vocabulary of predicates has been determined by the premise parsing step). This is in particular important when using a prompt-based LLM for semantic parsing.
After parsing into a ProbLog program, the solver executes the code to determine the answer.
A full ProbLog example for \datasetname is provided in \autoref{sec:example_calculation}.
For ProbLog-FT, we use Mistral-7B due to its large context window of 32k tokens.

\subsection{Baselines}
As a trivial baseline, we report a system always predicting $50\%$.
The \textbf{Regression-FT} is a Llama2-7B model \citep{touvron2023llama} 
trained for regression with sigmoid output given all premises, evidences and the question. 
The input to the regression layer is the embedding of the last token.
Additionally, we fine-tune a Mistral-7B model on predicting the probability as text, e.g., \enquote{The probability is $p$} (\textbf{LLM-FT}).

\section{Experimental Evaluation}
In this section, we report our extensive evaluation of current state-of-the-art LLMs and our neurosymbolic model on \datasetname. All fine-tuning details and parameters are provided in \autoref{sec:training_params}.

\subsection{Evaluation Metrics} 
As \datasetname comprises two versions of the background premises, we can investigate the following research questions:
(1) Given numeric premises, evidences, and a question, can the model(s) correctly calculate the likelihood of events/states?
(2) In the case of linguistically specified uncertainty (WEP-based premises), can the model provide close estimates %
of the likelihood of events/states? %
For each model, we hence report the percentages of \textbf{correct} predictions,\footnote{We evaluate with a relative tolerance of $10^{-4}$ to compensate floating point errors.} \textbf{wrong} predictions, and \textbf{error} cases without a valid numeric answer (e.g., in case of invalid ProgLog programs or if an LLM refuses to answer).
RSME, computed as $\sqrt{\sum_{i=1}^n \frac{(p_i - \hat{p_i})^2}{n}}$ %
reports how far numeric estimates deviate from the ground truth.
We report two variations for handling error cases: %
RMSE\textsubscript{50\%} %
has a fallback to $50\%$ as default answer for any invalid model output. %
The rationale behind choosing $50\%$ as fallback value for RMSE\textsubscript{50\%} is that whenever a model refuses to answer or produces invalid output (e.g., erroneous Problog code), we can only make a random guess.
Since $50\%$ reflects an equal likelihood of something being the case or not, it is a natural choice for the fallback case.

RMSE\textsubscript{nonError} is computed only over valid, but not necessarily correct predictions. %
Note that RMSE\textsubscript{nonError} scores are not comparable across rows.
RMSE\textsubscript{50\%} does not directly report the quality of \textit{valid} predictions, i.e., where the model or Problog solver return actual numeric values.
To also judge the quality of the valid numeric predictions, RMSE\textsubscript{nonError} only takes instances into account for which there are valid predictions.
Since the amount of valid predictions heavily alters between the different models and approaches, this score does not necessarily refer to the same instances between the different models, but can be used to interpret how close the numeric output of a model is to the correct answer.

\begin{table*}[t]
    \centering
    \footnotesize
    \setlength{\tabcolsep}{6pt}
  \scalebox{1.0}{
    \begin{tabular}{cll | lll | ll}
    \toprule
     & \multirow{2}{*}{\textbf{Method}} & \multirow{2}{*}{\textbf{Model}} & \multicolumn{3}{c|}{\textbf{Response Metrics}} & \multicolumn{2}{c}{\textbf{RMSE $\downarrow$}} \\
     \cmidrule(lr){4-6} \cmidrule(lr){7-8}
     & & & \textbf{\% correct $\uparrow$} & \textbf{\% wrong $\downarrow$} & \textbf{\% error$\downarrow$} %
        & \textbf{50\%} & \textbf{nonError}\\
    \midrule
    & \textit{$50\%$ baseline} & - & 0.9 & 99.1 & 0.0 %
        & 0.363 & 0.363\\
    & \textit{QE only baseline} & $_{\text{GPT4-Turbo}}$ & 3.1 & 96.9 & 0.0 %
        & 0.361 & 0.361 \\
    \midrule
    \multirow{11}{*}{\rotatebox[origin=c]{90}{\textit{\textbf{Numeric premises}}}}
    & \multirow{3}{*}{\textsc{Zero-Shot}} & $_{\text{GPT4-Turbo}}$ & 36.7 & 57.6 & 5.7 %
        & 0.304 & 0.302 \\
    & & $_{\text{Llama-3-8B}}$ & 7.0 & 91.7 & 1.3 %
        & 0.521 & 0.514 \\
    & & $_{\text{Mixtral-8x7B}}$ & 9.6 & 77.3 & 13.1 %
        & 0.441 & 0.449 \\
    \cmidrule(l){2-8}
    & \multirow{3}{*}{\causalcot} & $_{\text{GPT4-Turbo}}$ & 37.1 & 61.6 & 1.3 %
        & 0.326 & 0.326 \\
    & & $_{\text{Llama-3-8B}}$ & 11.4 & 87.8 & \textbf{0.9} %
        & 0.403 & 0.404 \\
    & & $_{\text{Mixtral-8x7B}}$ & 7.0 & 83.0 & 10.0 %
        & 0.486 & 0.498\\
    \cmidrule(l){2-8}
    & \textit{Regression-FT} & $_{\text{Llama-2-7B}}$ & 0.0$_{\pm 0.0}$ & 100.0$_{\pm 0.0}$ & 0.0$_{\pm 0.0}$ %
        & 0.369$_{\pm 0.00}$ & 0.369$_{\pm 0.00}$ \\
    & \textit{LLM-FT} & $_{\text{Mistral-7B}}$ & 21.5$_{\pm 1.4}$ & 78.5$_{\pm 1.4}$ & 0.0$_{\pm 0.0}$ %
        & 0.327$_{\pm 0.01}$ & 0.327$_{\pm 0.01}$ \\
    & \textit{ProbLog-Prompt} & $_{\text{GPT4-Turbo}}$ & 19.2 & \textbf{4.8} & 76.1 %
        & 0.313 & \textbf{0.116} \\
    & \textit{ProbLog-FT} & $_{\text{Mistral-7B}}$ & \textbf{54.5}$_{\pm 4.8}$ & 16.9$_{\pm 5.1}$ & 28.6$_{\pm 7.1}$ %
        & \textbf{0.244$_{\pm 0.03}$} & 0.179$_{\pm 0.05}$ \\
        \toprule
    \multirow{11}{*}{\rotatebox[origin=c]{90}{\textit{\textbf{WEP-based premises}}}}
    & \multirow{3}{*}{\textsc{Zero-Shot}} & $_{\text{GPT4-Turbo}}$  & 5.7 & 82.1 & 12.2 %
        & 0.391 & 0.394 \\
    & & $_{\text{Llama-3-8B}}$  & 2.2 & 83.4 & 14.4 %
        & 0.493 & 0.512 \\
    & & $_{\text{Mixtral-8x7B}}$  & 3.5 & 50.7 & 45.9 %
        & 0.484 & 0.562 \\
    \cmidrule(l){2-8}
    & \multirow{3}{*}{\causalcot} & $_{\text{GPT4-Turbo}}$  & 8.7 & 89.1 & 2.2 %
        & 0.341 & 0.341 \\
    & & $_{\text{Llama-3-8B}}$  & 3.5 & 91.7 & 4.8 %
        & 0.436 & 0.438 \\
    & & $_{\text{Mixtral-8x7B}}$ & 2.6 & 59.4 & 38.0 %
        & 0.456 & 0.511 \\
    \cmidrule(l){2-8}
    & \textit{Regression-FT} & $_{\text{Llama-2-7B}}$  & 0.0$_{\pm 0.0}$ & 100.0$_{\pm 0.0}$ & 0.0$_{\pm 0.0}$ %
        & 0.425$_{\pm 0.06}$ & 0.425$_{\pm 0.06}$ \\
    & \textit{LLM-FT} & $_{\text{Mistral-7B}}$ & 3.6$_{\pm 0.9}$ & 96.4$_{\pm 0.9}$ & 0.0$_{\pm 0.0}$ %
        & 0.344$_{\pm 0.01}$ & 0.344$_{\pm 0.01}$ \\
    & \textit{ProbLog-Prompt} & $_{\text{GPT4-Turbo}}$ & 0.4 \hspace{0.6cm} & 8.7 \hspace{0.6cm} & 90.8 %
        & 0.362 & \textbf{0.268}\\
    & \textit{ProbLog-FT} & $_{\text{Mistral-7B}}$ & 1.3$_{\pm 0.6}$ & 32.8$_{\pm 4.6}$ & 65.9$_{\pm 4.8}$ %
        & \textbf{0.319}$_{\pm 0.01}$ & 0.299$_{\pm 0.04}$ \\
    \midrule
    \midrule
    & \textit{ProbLog-FT oracle} &  $_{\text{Mistral-7B}}$ & 87.3$_{\pm 3.0}$ & 5.8$_{\pm 1.7}$ & 6.9$_{\pm 1.6}$ %
        & 0.165$_{\pm 0.04}$ & 0.145$_{\pm 0.04}$\\
    \bottomrule
    \end{tabular}
    }
    \caption{Results on \datasetname~for numeric and words of estimative probability (WEP)-based background premises.}%
    \label{tab:results_numeric}
    \vspace{-0.2cm}
\end{table*}

\subsection{Results for Numeric Premises}
The results for numeric background premises are reported in the upper part of Table~\ref{tab:results_numeric}. 
The QE-only baseline represents GPT-4 performance when omitting the premises.
Only 3.1\% of the cases are solvable by GPT-4 without referring to information in the premises, which means that background knowledge and reasoning is required for solving \datasetname instances.
This experiment validates the suitability of \datasetname to test for probabilistic reasoning performance. %

The Regression-FT model does not predict any instances correctly and has a similar performance as the baseline that always predicts the average probability value. %
LLM-FT correctly predicts the output in 1/5 of the cases without invalid responses, which implies that this method leveraged pre-training information in a better way.%

Next, we compare results for two prompting techniques. 
In terms of accuracy and RMSE, GPT-4 outperforms the open-weights models by a large margin.
\citet{jin2023cladder} report considerable performance gains (10-20\%) over zero-shot settings when using \causalcot on \cladder.
On \datasetname, we observe that both GPT-4 and Llama-3 slightly profit from the \causalcot technique, but the gain is not as large as observed on \cladder.
Results are somewhat inconclusive as performance for Mixtral even drops when integrating \causalcot.

ProbLog-FT outperforms all other models and approaches by a large margin. It is the only approach that finds the correct answer to about every second question. %
Outsourcing all mathematical steps that are required to obtain the final answer to an external solver is much more effective overall, also achieving the best RMSE\textsubscript{50} score, demonstrating a clear benefit over the trivial baselines and over all prompting-based approaches.
To substantiate the need for fine-tuning, we prompt GPT-4 on the task of generating ProbLog code (ProbLog-Prompt). %
This model %
suffers from producing a very high number of parsing errors (approximately 76\% of the cases).

\subsection{Results for WEP-based Premises}
In the case of WEP-based background premises (lower part of \autoref{tab:results_numeric}), we focus on the RMSE scores. %
Interestingly, when provided with the WEP-based premises, GPT-4 still arrives at the correct solution in $8.7\%$ of all cases, indicating that it leverages the textual descriptions in the premises.
Compared to using numeric premises, Problog-FT produces more parsing errors, indicating that more training data is necessary in this setting.
Most notably, however, ProbLog-FT (7B parameters) performs on par with GPT-4 (estimated 8x222B parameters), which indicates that fine-tuning neuro-symbolic models is a promising direction to improve automatic reasoning.

\subsection{Results by Reasoning Type}

\autoref{fig:reasoning_types} (on page 1) breaks down the results by reasoning type, showing how many instances in each category have been solved correctly by the best-performing models.
Causal reasoning seems to be the easiest type of reasoning.
All models except for ProbLog-FT fail on \textit{evidential} and \textit{explaining-away} reasoning.
We conclude that LLMs show reasonable skills in forward-style reasoning, whereas backward-style reasoning seems to be a major issue. %
Once a valid representation of the causal structure has been assembled, however, our neuro-symbolic models can perform any type of reasoning, illustrating an important advantage of our neuro-symbolic approach.

\subsection{Error Analysis}
In this section, we provide qualitative and quantitative analyses for different failure cases of our approaches. Further analyses on the effect of network size on performance are provided in \autoref{sec:further_analysis}.

\paragraph{Neuro-symbolic Approach.}
The ProbLog-FT model has two main sources of errors: syntax errors and unknown clauses, i.e., using undefined predicates.
To get a sense of whether the main source of errors is step 1 (premise parsing) or step 2 (QE parsing), we conduct an oracle experiment (cf. bottom row in \autoref{tab:results_numeric}) in which the already parsed premises are provided to the network and only QE parsing is performed by the model.
In this setup, the model is able to get four out of five cases right, which strongly indicates that parsing the lengthy premises is the main source of errors. %

\paragraph{Prompting.}
Our qualitative analysis reveals that for prompt-based LLM approaches, \textbf{mathematical errors} are a frequent error case, with wrong answers being produced due to rounding errors or erroneous calculations.
In other cases, the LLMs refuse to answer because of the mathematical complexity or asks whether it should continue. Occasionally, the models insist on not having enough information to solve the question and conclude the generation without results.

\section{Conclusion and Outlook}
In this paper, we have presented \datasetname, a new question answering dataset that provides Bayesian reasoning scenarios for a variety of domains and 
that can be used to assess uncertainty-based reasoning with LLMs.
From a large set of experiments using numeric probabilistic premises and premises expressed using words of estimative probability, we conclude that a neuro-symbolic approach combining probabilistic logic programming and fine-tuned LLMs as semantic parsers is most promising.
Moreover, we find that non-specialized LLMs mostly fail on this task. %

\paragraph{Outlook.}
For increasing the robustness of logic-based semantic parsing models, different approaches should be further investigated. For example, constrained decoding techniques could be used to ensure that only valid predicates can be generated at any point in time.
Next steps include also studying reasoning in modal and counterfactual scenarios.

\section*{Limitations}
In this work, we investigate probabilistic reasoning with uncertainty using verbalized Bayesian networks.
This approach assumes that the entire probability distribution is known and given at any time.
However, in real-life scenarios (e.g., in data that is obtained from production plants), probability distributions are often underspecified or influencing factors are not even known, i.e., there are hidden variables that influence the relationships.
Furthermore, our models act upon a limited number of sentences at once, whereas uncertainty descriptions could also be provided in longer texts that also contain information that is not relevant to the reasoning process.

In its current version, \datasetname operates on \textit{rung 1} of the \textit{ladder of causation} \citep{bookofwhy}. Our work could of course be extended in the future to also cover rung 2 of the \textit{ladder of causation}, meaning that based on the networks, one could perform interventional queries (\textit{do-operator}) that dynamically modify the probabilistic relationships.
To do that, we need to generate additional queries of form \textit{If we force [...], does that lead to [...]} and map that onto the \textit{do/1} predicate in ProbLog.
As a first step, however, we decided to carefully study rung-1 questions with regard to three Bayesian reasoning types.

As with most benchmarks these days, there is a potential issue of data contamination, i.e., the LLMs could have seen relevant parts of \datasetname in their pre-training corpus.
Our natural language corpus is based on plain probability tables.
These tables could have been part of the pretraining corpus.
Some of the networks in our dataset were described in published work before.
Therefore, it could be that some of the relationships between random variables are vaguely known to the LLMs.
However, we argue that no paper describes large BNs in every detail, preventing the LLMs from learning every network detail by hard. This assumption is supported by the poor performance of the question-evidence only baseline.

\section*{Ethical Considerations}
\datasetname builds upon data from many different scientific and non-scientific domains.
These include different Bayesian networks from domains related to medical treatment and health issues.
However, we emphasize that \datasetname and our proposed models in their current version should not be used for any kind of reliable decision making in medicine and health-related issues.
All probabilistic networks in \datasetname only reflect a subset of the entire causal relationships that might exist and are used for assessing self-contained Bayesian reasoning without considering the much broader scientific knowledge available to the world.
Furthermore, we did not verify the correctness of the observed data by checking the biomedical literature.

\section*{Acknowledgements}
We would like to express our deepest gratitude to Marco Scutari, who is the author of \textit{bnlearn} and gave us the permission to use the Bayesian networks for our research.
We also thank the anonymous reviewers for their valuable feedback.
Finally, we would like to thank our colleagues at Bosch Research for all the valuable discussions, feedback and ideas on and for our work.

\bibliography{custom}

\appendix

\newpage

\section{Theoretical Background}
\label{sec:theoretical_background}
This section introduces the main theoretical concepts on probabilistic reasoning. We closely follow the notations and definitions of \citet{koller2009probabilistic}.
\subsection{Bayesian Networks}
A Bayesian network is a directed acyclic graph (DAG) $\mathcal{G} = (\mathbf{V}, \mathbf{E})$ that represents a joint probability distribution $\mathbb{P}$ over a set of random variables $\{X_1, \dots, X_n\}$.
Each random variable $X_i$ can take values from their respective domain $\Omega_i$, which is the set of possible realizations $\{x_i^1,\dots,x_i^k\}$, where $k = |\Omega_i|$.
If $k = 2$, then we say that $X_i$ adheres to a \textit{Bernoulli} distribution, whereas $k > 2$ makes them a \textit{categorical} random variable.
Furthermore, to ensure a valid probability distribution, it must always hold that $\sum_{j=1}^{k} \mathbb{P}(X_i = x_i^j) = 1$.

Each node $v_i \in \mathbf{V}$ in $\mathcal{G}$ represents one variable $X_i$.
An edge $e_{i,j} \in \mathbf{E}$ between two nodes $v_i$, $v_j$ represents a correlation between the two associated random variables and thereby models the conditional probability distribution (CPD) $\mathbb{P}(X_j | X_i)$.
These CPDs are of special importance when dealing with so-called \textit{observations}.
Mathematically speaking, observations modify the joint probability distribution over $\mathcal{G}$ as follows:
\begin{align*}
    \mathbb{P}(X_1,\dots,X_{j-1},X_{j+1},\dots,X_{n}|X_j = x_j^{m}) \\ = \frac{ \mathbb{P}(X_1,\dots,X_j = x_j^{m}\dots,X_{n})}{\mathbb{P}(X_j = x_j^{m})}
\end{align*}
This means, to obtain the value of the full probability distribution, only one state of $X_j$ must be considered instead of the whole range of possible states.

\subsection{Reasoning Patterns}
\label{ssec:reasoning_patterns}
Directed edges in a Bayesian network not only indicate a dependence, but they also allow for different types of reasoning when observing one or more random variables.
In the following, we assume a simple three-node network with nodes $X_1, X_2, X_3$ and edges $X_1 \xrightarrow{} X_3$ and $X_2 \xrightarrow{} X_3$.
\begin{description}
    \item[Causal Reasoning:] We know the cause and draw conclusions about the effect. Suppose we look at the connection $X_1 \xrightarrow{} X_3$ and observe the value of $X_1$.
    This gives us a strong hint on what the status of $X_3$ could likely be.
    The underlying probability distribution is $\mathbb{P}(X_3|X_1)$.
    For instance, assume we observe that the weather is rainy. This leads us to the conclusion that the streets are very likely to be wet.
    \item[Evidential Reasoning:] The other way round is to observe $X_3$ and reason about $X_1$ in $X_1 \xrightarrow{} X_3$.
    We now know about the value of the effect, which we can use to make assumptions about the most likely cause.
    The mathematical term is $\mathbb{P}(X_1|X_3)$.
    This probability is not directly represented in this Bayesian network, but it can be calculated by using Bayes' theorem: $\mathbb{P}(X_1|X_3) = \frac{\mathbb{P}(X_1, X_3)}{\mathbb{P}(X_3)}$.
    Let us now assume we observe wet streets. This gives us strong hints on whether it has been raining before.
    \item [Explaining-Away:] This requires multiple causes with a common effect, shaping a so-called \textit{v-structure} ($X_1 \xrightarrow{} X_3 \xleftarrow{} X_2$). 
    Now observing one of the potential causes ($X_1$ or $X_2$) and the effect \enquote{explains the influence of the other causes away.}
    Assume that we again observe wet streets. This could be due to rain, but also due road cleaning machines. If we now obtain the knowledge that it is raining or was raining, we can assume that road cleaning is unlikely. The state of the weather \enquote{explains away} the need for road cleaning machines.
\end{description}

We enrich our dataset by categorizing the queries into their respective reasoning type(s) if applicable, making it possible to also investigate the robustness of LLMs with respect to different reasoning patterns.

\section{Words of Estimative Probability (WEP)}
\label{sec:wep}
\autoref{tab:wep} lists the WEPs which are used to model uncertainty in \datasetname.
The table provides a mapping between adverbs and numeric probabilities, estimated via a survey conducted by \citet{fagen}.
Every numeric value is mapped to the closest adverb, not considering the confidence intervals in the table.
However, we introduce one exception in the mapping: if a numeric probability is below $45\%$, it is not mapped to the closest adverb (i.e., \textit{about even}), but instead to \textit{probably not}. 
This is to make sure that values of $38\%$ for instance are not mapped to \textit{about even}.

\begin{table}[!h]
    \centering
    \footnotesize
    \begin{tabular}{ll}
    \toprule
    \textbf{WEP} & \textbf{Associated Prob.} \\
    \midrule
    \textit{certain} & $100\%$ \\
    \textit{almost certain} & $95.0\% \pm 10.9\%$ \\
    \textit{highly likely} & $ 90.0\% \pm 8.4\%$\\
    \textit{very good chance} & $ 80.0\% \pm 10.8\%$\\
    \textit{likely} & $ 70.0\% \pm 11.3\%$\\
    \textit{probably} & $ 70.0\% \pm 12.9\%$\\
    \textit{probable} & $ 70.0\% \pm 14.7\%$\\
    \textit{better than even} & $ 60.0\% \pm 9.1\%$\\
    \textit{about even} & $ 50.0\% \pm 4.9\%$\\
    \textit{probably not} & $ 25.0\% \pm 14.4\%$\\
    \textit{unlikely} & $ 20.0\% \pm 15.0\%$\\
    \textit{little chance} & $ 10.0\% \pm 12.2\%$\\
    \textit{chances are slight} & $10.0\% \pm 10.9\%$\\
    \textit{improbable} & $10.0\% \pm 17.5\%$\\
    \textit{highly unlikely} & $5.0\% \pm 17.3\%$\\
    \textit{almost no chance} & $2.0\% \pm 17.0\%$\\
    \textit{impossible} & $0\%$ \\
    \bottomrule
    \end{tabular}
    \caption{The mapping of WEP to numeric probabilities by \citet{fagen} that we use to model uncertainty on the Bayesian networks in our dataset.}
    \label{tab:wep}
\end{table}

\section{Fine-Tuning Details}
\label{sec:training_params}
We fine-tune different modelling approaches on \datasetname as described in Section \ref{sec:modelling} to see if specialized models can outperform out-of-the-box baseline models.
Since state-of-the-art models are made up of billions of parameters, we use LoRA \citep{hu2021lora} to fine-tune low-rank adapters on \datasetname.
We use the code from \citet{alignment_handbook2023} to fine-tune all models on Nvidia H100 and A100 GPUs.
All models are trained using full precision, i.e., FP32 to make sure that we do not lose performance.
The hyperparameters for each model are listed in \autoref{tab:train_params}.
To keep fine-tuning sustainable, we refrained from performing an extensive hyperparameter search.
Instead, we selected commonly used values (e.g., cf. \citet{zheng2024llamafactory}).
Models are selected based on their performance on the development set, which is a subset of train, and evaluated only once on test.

\begin{table}[t]
    \centering
    \footnotesize
    \begin{tabular}{l|ccc}
    \toprule
    \textbf{Hyp.-param.} & \textbf{ProbLog-FT} & \textbf{Regr.-FT} & \textbf{TP-FT}\\
    \midrule
    \textit{Learning rate} & $5e-5$ & $5e-5$ & $5e-5$ \\
    \textit{LoRA rank} & 64 & 64 & 64\\
    \textit{LoRA $\alpha$} & 32 & 32 & 32\\
    \textit{Batch size} & 4 & 4 & 4\\
    \textit{Epochs} & 12 & 10 & 10\\
    \midrule
    \textit{Num. GPUs} & 4 & 1 & 1\\
    \bottomrule
    \end{tabular}
    \caption{Parameters for LoRA-based fine-tuning on \datasetname.}
    \label{tab:train_params}
\end{table}

\section{Exemplary Probability Calculation and Reasoning Steps}
\label{sec:example_calculation}
\begin{figure*}[t]
    \centering
    \includegraphics[width=1.0\linewidth]{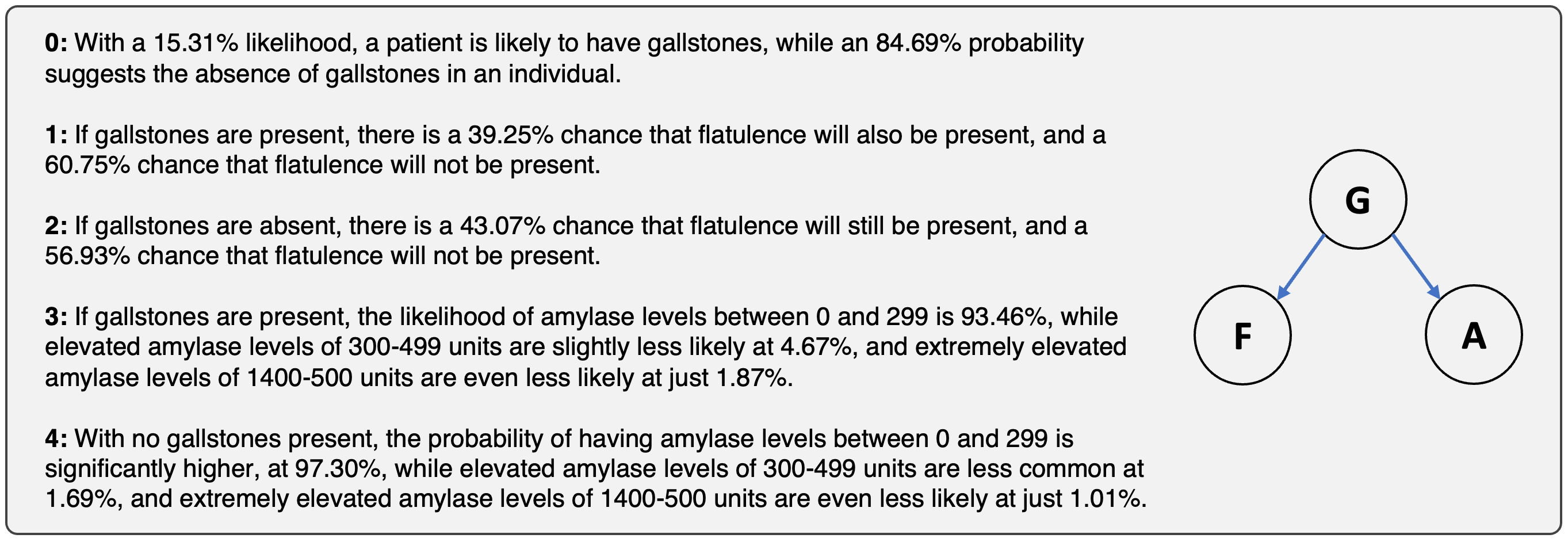}
    \caption{Exemplary network from \datasetname about the relationship between gallstones, flatulence and amylase levels.}
    \label{fig:example_bayes_calculation}
\end{figure*}

We now take a look at the smallest network instance in \datasetname as shown in \autoref{fig:example_bayes_calculation} and manually execute the reasoning steps. The network is about the relationship of gallstones, flatulence, and amylase levels.
We have three random variables here which we are going to refer to as $G$ (gallstones), $F$ (flatulence), and $A$ (amylase levels).
It is important to mention that the models are only given the background premises, i.e., statements 0, 1, 2, 3, and 4.
In a first step, they have to build up the graph shown, either as internal representations in their embeddings, as textual output in a chain-of-thought style of reasoning, or explicitly as ProbLog code.
Furthermore, it is up to the model to determine not only the involved random variables, but also their possible states, i.e., $\Omega_i$.
For gallstones ($G$) and flatulence ($F$), it is a \enquote{yes} or \enquote{no} decision, i.e., $|\Omega_G| = |\Omega_F| = |\{yes, no \}| = 2$.
Amylase levels show that \datasetname also contains many categorical variables since $A$ can take the values \enquote{0-299}, \enquote{300-499}, or \enquote{500-1400}, i.e., $|\Omega_A| = |\{0-299, 300-499, 500-1400 \}| = 3$.

Next, we look at a specific question-evidence (QE) pair with one observiation (evidence) and one question:
\begin{itemize}
    \item \textbf{Evidence}: \textit{The patient has flatulence.}
    \item \textbf{Question}: \textit{What is the likelihood of a patient having an amylase level between 1400 and 500 U/L?}
\end{itemize}

Before we can start calculating the answer to the question, we first need to understand the joint probability distribution that is represented by the network.
It is a joint probability distribution over three random variables.
Furthermore, the chain rule allows us to break down the joint probability into its single components:

\begin{gather*}
    \mathbb{P}(G, F, A) = \mathbb{P}(F|G) \cdot \mathbb{P}(A|G) \cdot \mathbb{P}(G)
\end{gather*}

We now calculate:

\begin{gather*}
    \mathbb{P}(A = 500-1400 | F = yes)
\end{gather*}

According to the Bayes' theorem, this is equivalent to the joint probability over both variables divided by the condition:

\begin{gather*}
    \mathbb{P}(A = 500-1400 | F = yes) \\ = \frac{\mathbb{P}(A = 500-1400, F = yes)}{\mathbb{P}(F = yes)}
\end{gather*}

However, neither probability is explicitly given in the network.
Instead, we only have conditional probabilities or the joint probability over all three random variables, not only two of them.
To obtain the necessary probabilities, we need to \enquote{marginalize out} unwanted random variables by summing over all states:

\begin{gather*}
    \mathbb{P}(A = 500-1400, F = yes) 
    \\ = \sum_G \mathbb{P}(A = 500-1400, F = yes, G)
    \\ = \mathbb{P}(A = 500-1400, F = yes, G = yes)
    \\ + \mathbb{P}(A = 500-1400, F = yes, G = no)
\end{gather*}

We obtain $\mathbb{P}(A = 500-1400, F = yes, G = yes)$ by using the single conditional probabilities stated in the network:

\begin{gather*}
    \mathbb{P}(A = 500-1400, F = yes, G = yes)
    \\ = \mathbb{P}(F=yes|G=yes)
    \\ \cdot \mathbb{P}(A=500-1400|G=yes)
    \\ \cdot \mathbb{P}(G=yes)
    \\ = 0.3925 \cdot 0.0187 \cdot 0.1531 \approx 0.001124
\end{gather*}

Doing the same for $G = no$ yields:

\begin{gather*}
    \mathbb{P}(A = 500-1400, F = yes, G = no)
    \\ = 0.4307 \cdot 0.0101 \cdot 0.8469 \approx 0.003684
\end{gather*}

$\mathbb{P}(A = 500-1400, F = yes)$ is now the sum of both parts, i.e.

\begin{gather*}
    \mathbb{P}(A = 500-1400, F = yes) \approx 0.004808
\end{gather*}

Obtaining $\mathbb{P}(F = yes)$ is as straightforward as just shown, but requires two marginalization steps, making it more calculation work:

\begin{gather*}
    \mathbb{P}(F = yes) 
    \\ = \sum_{G, A} \mathbb{P}(F = yes, G, A)
\end{gather*}

Repeating the same steps from above, but with 6 summands since $G$ can take 2 states and $A$ 3 states yields the following probability:

\begin{gather*}
    \mathbb{P}(F = yes) 
    \\ = \sum_{G, A} \mathbb{P}(F = yes, G, A)
    \\ = \mathbb{P}(F = yes, G = yes, A = 0-299)
    \\ + \mathbb{P}(F = yes, G = yes, A = 300-499)
    \\ + \mathbb{P}(F = yes, G = yes, A = 500-1400)
    \\ + \mathbb{P}(F = yes, G = no, A = 0-299)
    \\ + \mathbb{P}(F = yes, G = no, A = 300-499)
    \\ + \mathbb{P}(F = yes, G = no, A = 500-1400)
    \\ \approx 0.424856
\end{gather*}

Finally, we can obtain the answer to the question:

\begin{gather*}
    \mathbb{P}(A = 500-1400 | F = yes) = \frac{0.004808}{0.424856} 
    \\ \approx \mathbf{0.01132}
\end{gather*}

Therefore, we conclude that the probability of having amylase levels between 500 and 1400 given the presence of flatulence is approximately $1.132\%$.

The models are now expected to either perform this calculation by themselves (either explicitly or implicitly) or parse the entire problem into ProbLog representation such that the ProbLog executable can calculate the actual answer.

\begin{figure}
    \centering
    \begin{lstlisting}
0.1531::gallstones(patient).

0.3925::flatulence(patient) :- gallstones(patient).

0.4307::flatulence(patient) :- not gallstones(patient).

0.9346::amylase(patient, '0-299'); 0.0467::amylase(patient, '300-499'); 0.0187::amylase(patient, '500-1400') :- gallstones(patient).

0.9730::amylase(patient, '0-299'); 0.0169::amylase(patient, '300-499'); 0.0101::amylase(patient, '500-1400') :- not gallstones(patient).

evidence(flatulence(patient), true).

query(amylase(patient, '500-1400')).

\end{lstlisting}
    \caption{Full ProbLog code for the gallstone-flatulence-amylase instance.}
    \label{fig:problog_code_full}
\end{figure}

\autoref{fig:problog_code_full} depicts how this mathematical problem is represented in ProbLog.
The first five statements define the probabilistic model.
We expect the model to perform semantic parsing from the natural language input to this structured representation.
Each right-hand side represents the conditions for the left-hand side.
Furthermore, since amylase levels is a categorical variable, it is necessary to connect all possible states via a semicolon in order to match them to the same probability distribution.
This is an additional difficulty of \datasetname since models can not rely on just writing down binary predicates.
Next, the \textit{evidence/2} predicate is used to set the observation.
Here it is of key importance that the model only reuses predicates that were already defined in the parsed premises above.
Finally, the question is defined using \textit{query/1}.
When calling ProbLog on this program, it outputs \textit{amylase(patient,'500-1400'):    0.011316399}, which perfectly matches our calculation by hand.

\section{BN Subsetting}
\label{sec:bn_subsetting}
\begin{figure}
    \centering
    \includegraphics[width=0.5\linewidth]{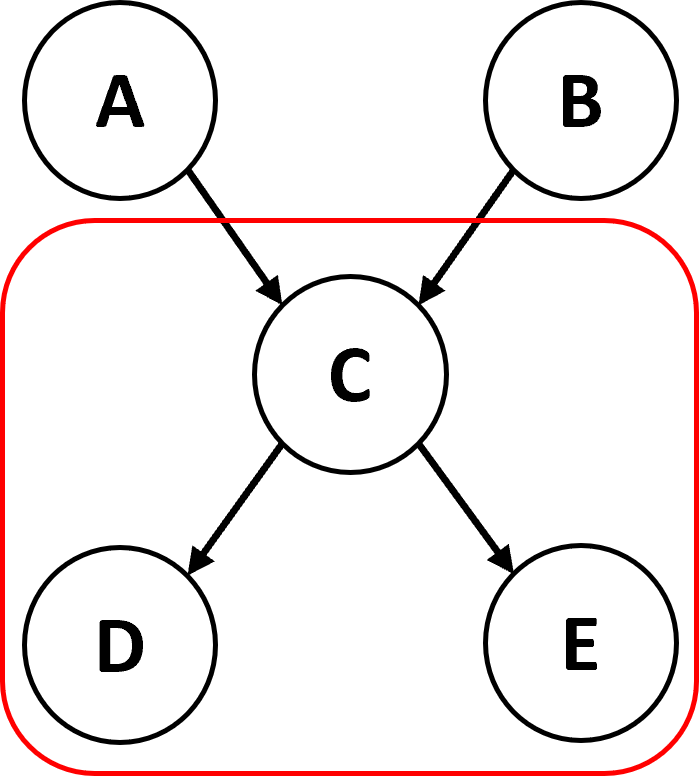}
    \caption{Example network for demonstrating our procedure of subsetting.}
    \label{fig:subset_network}
\end{figure}

Assume we want to extract the network $D \xleftarrow{} C \xrightarrow{} E$ from the five-node network depicted in \autoref{fig:subset_network}.
It is not possible to just cut the connections $A \xrightarrow{} C$ and $B \xrightarrow{} C$ since node $C$ only holds tables for CPDs that depend on $A$ and $B$ respectively.
Therefore, we modify the subnetwork by marginalizing out $A$ and $B$ from the probability distribution of $C$:

\begin{gather*}
    \mathbb{P}(C) = \sum_{A,B} \mathbb{P}(C, A, B) 
    \\ = \sum_{A,B} \mathbb{P}(C|A) \cdot \mathbb{P}(C|B) \cdot \mathbb{P}(A) \cdot \mathbb{P}(B)
\end{gather*}

\section{Further Dataset Statistics}

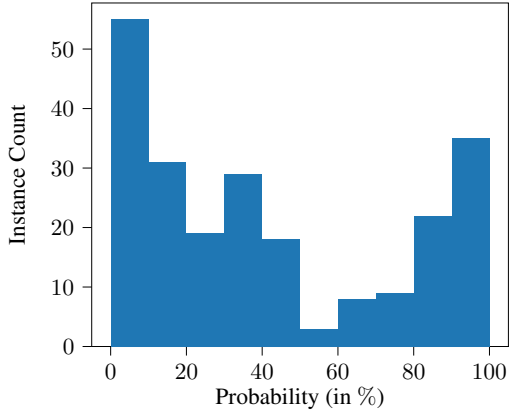
\begin{figure}
    \centering
    \scalebox{0.8}{\begin{tikzpicture}

\definecolor{darkgray176}{RGB}{176,176,176}
\definecolor{steelblue31119180}{RGB}{31,119,180}

\begin{axis}[
tick align=outside,
tick pos=left,
x grid style={darkgray176},
xmin=-5, xmax=105,
xtick style={color=black},
y grid style={darkgray176},
ymin=0, ymax=57.75,
ytick style={color=black},
xlabel={Probability (in $\%$)},
ylabel={Instance Count},
]
\draw[draw=none,fill=steelblue31119180] (axis cs:0,0) rectangle (axis cs:10,55);
\draw[draw=none,fill=steelblue31119180] (axis cs:10,0) rectangle (axis cs:20,31);
\draw[draw=none,fill=steelblue31119180] (axis cs:20,0) rectangle (axis cs:30,19);
\draw[draw=none,fill=steelblue31119180] (axis cs:30,0) rectangle (axis cs:40,29);
\draw[draw=none,fill=steelblue31119180] (axis cs:40,0) rectangle (axis cs:50,18);
\draw[draw=none,fill=steelblue31119180] (axis cs:50,0) rectangle (axis cs:60,3);
\draw[draw=none,fill=steelblue31119180] (axis cs:60,0) rectangle (axis cs:70,8);
\draw[draw=none,fill=steelblue31119180] (axis cs:70,0) rectangle (axis cs:80,9);
\draw[draw=none,fill=steelblue31119180] (axis cs:80,0) rectangle (axis cs:90,22);
\draw[draw=none,fill=steelblue31119180] (axis cs:90,0) rectangle (axis cs:100,35);
\end{axis}

\end{tikzpicture}}
    \caption{Distribution of probability values of the QE pairs in the test set of \datasetname.}
    \label{fig:prob_dist}
\end{figure}

\begin{figure}[t]
    \centering
    \scalebox{0.8}{\begin{tikzpicture}

\definecolor{darkgray176}{RGB}{176,176,176}
\definecolor{steelblue31119180}{RGB}{31,119,180}

\begin{axis}[
tick align=outside,
tick pos=left,
x grid style={darkgray176},
xmin=-0.89, xmax=29.89,
xtick style={color=black},
y grid style={darkgray176},
ymin=0, ymax=13.65,
ytick style={color=black},
xlabel={Network Number},
ylabel={Node Count},
xticklabel=\empty,
grid
]
\draw[draw=none,fill=steelblue31119180] (axis cs:-0.4,0) rectangle (axis cs:0.4,3);
\draw[draw=none,fill=steelblue31119180] (axis cs:0.6,0) rectangle (axis cs:1.4,3);
\draw[draw=none,fill=steelblue31119180] (axis cs:1.6,0) rectangle (axis cs:2.4,3);
\draw[draw=none,fill=steelblue31119180] (axis cs:2.6,0) rectangle (axis cs:3.4,3);
\draw[draw=none,fill=steelblue31119180] (axis cs:3.6,0) rectangle (axis cs:4.4,3);
\draw[draw=none,fill=steelblue31119180] (axis cs:4.6,0) rectangle (axis cs:5.4,4);
\draw[draw=none,fill=steelblue31119180] (axis cs:5.6,0) rectangle (axis cs:6.4,4);
\draw[draw=none,fill=steelblue31119180] (axis cs:6.6,0) rectangle (axis cs:7.4,4);
\draw[draw=none,fill=steelblue31119180] (axis cs:7.6,0) rectangle (axis cs:8.4,4);
\draw[draw=none,fill=steelblue31119180] (axis cs:8.6,0) rectangle (axis cs:9.4,4);
\draw[draw=none,fill=steelblue31119180] (axis cs:9.6,0) rectangle (axis cs:10.4,5);
\draw[draw=none,fill=steelblue31119180] (axis cs:10.6,0) rectangle (axis cs:11.4,5);
\draw[draw=none,fill=steelblue31119180] (axis cs:11.6,0) rectangle (axis cs:12.4,5);
\draw[draw=none,fill=steelblue31119180] (axis cs:12.6,0) rectangle (axis cs:13.4,6);
\draw[draw=none,fill=steelblue31119180] (axis cs:13.6,0) rectangle (axis cs:14.4,6);
\draw[draw=none,fill=steelblue31119180] (axis cs:14.6,0) rectangle (axis cs:15.4,6);
\draw[draw=none,fill=steelblue31119180] (axis cs:15.6,0) rectangle (axis cs:16.4,6);
\draw[draw=none,fill=steelblue31119180] (axis cs:16.6,0) rectangle (axis cs:17.4,6);
\draw[draw=none,fill=steelblue31119180] (axis cs:17.6,0) rectangle (axis cs:18.4,6);
\draw[draw=none,fill=steelblue31119180] (axis cs:18.6,0) rectangle (axis cs:19.4,6);
\draw[draw=none,fill=steelblue31119180] (axis cs:19.6,0) rectangle (axis cs:20.4,7);
\draw[draw=none,fill=steelblue31119180] (axis cs:20.6,0) rectangle (axis cs:21.4,7);
\draw[draw=none,fill=steelblue31119180] (axis cs:21.6,0) rectangle (axis cs:22.4,7);
\draw[draw=none,fill=steelblue31119180] (axis cs:22.6,0) rectangle (axis cs:23.4,7);
\draw[draw=none,fill=steelblue31119180] (axis cs:23.6,0) rectangle (axis cs:24.4,8);
\draw[draw=none,fill=steelblue31119180] (axis cs:24.6,0) rectangle (axis cs:25.4,8);
\draw[draw=none,fill=steelblue31119180] (axis cs:25.6,0) rectangle (axis cs:26.4,8);
\draw[draw=none,fill=steelblue31119180] (axis cs:26.6,0) rectangle (axis cs:27.4,9);
\draw[draw=none,fill=steelblue31119180] (axis cs:27.6,0) rectangle (axis cs:28.4,12);
\draw[draw=none,fill=steelblue31119180] (axis cs:28.6,0) rectangle (axis cs:29.4,13);
\end{axis}

\end{tikzpicture}}
    \caption{Node count for each of the 30 networks in \datasetname.}
    \label{fig:network_size}
\end{figure}
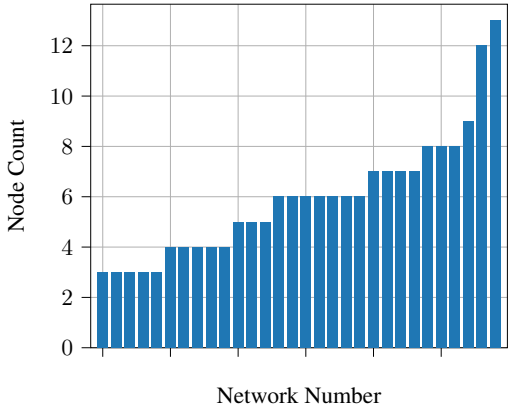
\begin{figure}[t]
    \centering
    \scalebox{0.8}{\begin{tikzpicture}

\definecolor{darkgray176}{RGB}{176,176,176}
\definecolor{steelblue31119180}{RGB}{31,119,180}

\begin{axis}[
tick align=outside,
tick pos=left,
x grid style={darkgray176},
xmin=-1.89, xmax=29.89,
xtick style={color=black},
y grid style={darkgray176},
ymin=0, ymax=124.95,
ytick style={color=black},
ylabel={Premise Count},
xlabel={Network Number},
xticklabel=\empty,
grid
]
\draw[draw=none,fill=steelblue31119180] (axis cs:-0.4,0) rectangle (axis cs:0.4,5);
\draw[draw=none,fill=steelblue31119180] (axis cs:0.6,0) rectangle (axis cs:1.4,8);
\draw[draw=none,fill=steelblue31119180] (axis cs:1.6,0) rectangle (axis cs:2.4,9);
\draw[draw=none,fill=steelblue31119180] (axis cs:2.6,0) rectangle (axis cs:3.4,9);
\draw[draw=none,fill=steelblue31119180] (axis cs:3.6,0) rectangle (axis cs:4.4,10);
\draw[draw=none,fill=steelblue31119180] (axis cs:4.6,0) rectangle (axis cs:5.4,13);
\draw[draw=none,fill=steelblue31119180] (axis cs:5.6,0) rectangle (axis cs:6.4,14);
\draw[draw=none,fill=steelblue31119180] (axis cs:6.6,0) rectangle (axis cs:7.4,16);
\draw[draw=none,fill=steelblue31119180] (axis cs:7.6,0) rectangle (axis cs:8.4,18);
\draw[draw=none,fill=steelblue31119180] (axis cs:8.6,0) rectangle (axis cs:9.4,18);
\draw[draw=none,fill=steelblue31119180] (axis cs:9.6,0) rectangle (axis cs:10.4,18);
\draw[draw=none,fill=steelblue31119180] (axis cs:10.6,0) rectangle (axis cs:11.4,29);
\draw[draw=none,fill=steelblue31119180] (axis cs:11.6,0) rectangle (axis cs:12.4,29);
\draw[draw=none,fill=steelblue31119180] (axis cs:12.6,0) rectangle (axis cs:13.4,29);
\draw[draw=none,fill=steelblue31119180] (axis cs:13.6,0) rectangle (axis cs:14.4,32);
\draw[draw=none,fill=steelblue31119180] (axis cs:14.6,0) rectangle (axis cs:15.4,36);
\draw[draw=none,fill=steelblue31119180] (axis cs:15.6,0) rectangle (axis cs:16.4,37);
\draw[draw=none,fill=steelblue31119180] (axis cs:16.6,0) rectangle (axis cs:17.4,39);
\draw[draw=none,fill=steelblue31119180] (axis cs:17.6,0) rectangle (axis cs:18.4,39);
\draw[draw=none,fill=steelblue31119180] (axis cs:18.6,0) rectangle (axis cs:19.4,44);
\draw[draw=none,fill=steelblue31119180] (axis cs:19.6,0) rectangle (axis cs:20.4,45);
\draw[draw=none,fill=steelblue31119180] (axis cs:20.6,0) rectangle (axis cs:21.4,50);
\draw[draw=none,fill=steelblue31119180] (axis cs:21.6,0) rectangle (axis cs:22.4,52);
\draw[draw=none,fill=steelblue31119180] (axis cs:22.6,0) rectangle (axis cs:23.4,67);
\draw[draw=none,fill=steelblue31119180] (axis cs:23.6,0) rectangle (axis cs:24.4,67);
\draw[draw=none,fill=steelblue31119180] (axis cs:24.6,0) rectangle (axis cs:25.4,71);
\draw[draw=none,fill=steelblue31119180] (axis cs:25.6,0) rectangle (axis cs:26.4,76);
\draw[draw=none,fill=steelblue31119180] (axis cs:26.6,0) rectangle (axis cs:27.4,87);
\draw[draw=none,fill=steelblue31119180] (axis cs:27.6,0) rectangle (axis cs:28.4,117);
\draw[draw=none,fill=steelblue31119180] (axis cs:28.6,0) rectangle (axis cs:29.4,119);
\end{axis}

\end{tikzpicture}}
    \caption{Premise count for each of the 30 networks in \datasetname.}
    \label{fig:premise_count}
\end{figure}
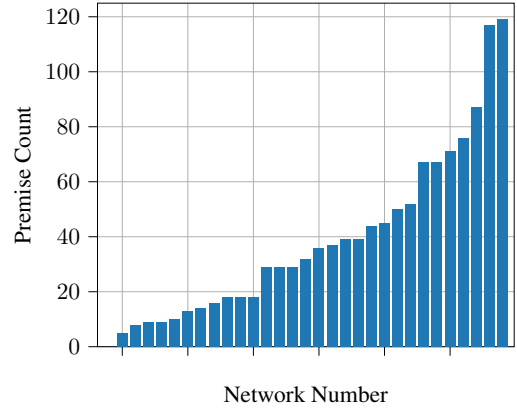

In \autoref{sec:example_calculation}, we used a three node network from \datasetname.
This was for demonstration purposes.
However, \datasetname contains networks of much larger sizes, i.e., with up to 13 nodes.
\autoref{fig:network_size} shows the distribution of node counts across the 30 networks in the dataset.
The largest network holds a joint probability distribution over 13 random variables.
Also, more than half of the networks do have at least 5 nodes.
The median size is 6, showing that the majority of our networks have of large size.

\autoref{fig:premise_count} depicts the distribution of premise counts in \datasetname. 
The median of the distribution is $34$ and the maximum is $119$.
This shows that building up the probabilistic model from the set of premises is already a computationally demanding task and requires very long context understanding.

\section{Further Analysis}
\label{sec:further_analysis}
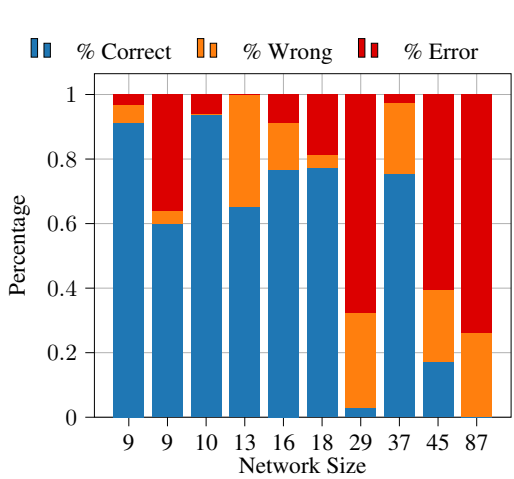
\begin{figure}
    \centering
    \scalebox{0.8}{\begin{tikzpicture}

\definecolor{darkgray176}{RGB}{176,176,176}
\definecolor{darkorange25512714}{RGB}{255,127,14}
\definecolor{redred}{RGB}{220, 0, 0}
\definecolor{lightgray204}{RGB}{204,204,204}
\definecolor{steelblue31119180}{RGB}{31,119,180}

\begin{customlegend}[legend columns=5,legend style={align=center,draw=none,column sep=2ex},
width=4.8in,
legend entries={\% Correct, \% Wrong, \% Error}]
\addlegendimage{ybar,ybar legend,draw=none,fill=steelblue31119180}
\addlegendimage{ybar,ybar legend,draw=none,fill=darkorange25512714}
\addlegendimage{ybar,ybar legend,draw=none,fill=redred}
\end{customlegend}
\end{tikzpicture}}

\scalebox{0.8}{\begin{tikzpicture}

\definecolor{darkgray176}{RGB}{176,176,176}
\definecolor{darkorange25512714}{RGB}{255,127,14}
\definecolor{redred}{RGB}{220, 0, 0}
\definecolor{lightgray204}{RGB}{204,204,204}
\definecolor{steelblue31119180}{RGB}{31,119,180}

\begin{axis}[
legend cell align={left},
legend style={fill opacity=0.8, draw opacity=1, text opacity=1, draw=lightgray204},
tick align=outside,
tick pos=left,
x grid style={darkgray176},
xmin=0.11, xmax=10.89,
xtick style={color=black},
y grid style={darkgray176},
ymin=0, ymax=1.06448275862069,
ytick style={color=black},
xtick={1,2,3,4,5,6,7,8,9,10},
xticklabels={9, 9, 10, 13, 16, 18, 29, 37, 45, 87},
grid,
xlabel={Network Size},
ylabel={Percentage}
]
\draw[draw=none,fill=steelblue31119180] (axis cs:0.6,0) rectangle (axis cs:1.4,0.911111111111111);
\draw[draw=none,fill=steelblue31119180] (axis cs:1.6,0) rectangle (axis cs:2.4,0.6);
\draw[draw=none,fill=steelblue31119180] (axis cs:2.6,0) rectangle (axis cs:3.4,0.938461538461538);
\draw[draw=none,fill=steelblue31119180] (axis cs:3.6,0) rectangle (axis cs:4.4,0.65);
\draw[draw=none,fill=steelblue31119180] (axis cs:4.6,0) rectangle (axis cs:5.4,0.76551724137931);
\draw[draw=none,fill=steelblue31119180] (axis cs:5.6,0) rectangle (axis cs:6.4,0.773333333333333);
\draw[draw=none,fill=steelblue31119180] (axis cs:6.6,0) rectangle (axis cs:7.4,0.0275862068965517);
\draw[draw=none,fill=steelblue31119180] (axis cs:7.6,0) rectangle (axis cs:8.4,0.754545454545455);
\draw[draw=none,fill=steelblue31119180] (axis cs:8.6,0) rectangle (axis cs:9.4,0.173333333333333);
\draw[draw=none,fill=steelblue31119180] (axis cs:9.6,0) rectangle (axis cs:10.4,0);
\draw[draw=none,fill=darkorange25512714] (axis cs:0.6,0.911111111111111) rectangle (axis cs:1.4,0.966666666666667);

\draw[draw=none,fill=darkorange25512714] (axis cs:1.6,0.6) rectangle (axis cs:2.4,0.64);
\draw[draw=none,fill=darkorange25512714] (axis cs:2.6,0.938461538461538) rectangle (axis cs:3.4,0.938461538461538);
\draw[draw=none,fill=darkorange25512714] (axis cs:3.6,0.65) rectangle (axis cs:4.4,1);
\draw[draw=none,fill=darkorange25512714] (axis cs:4.6,0.76551724137931) rectangle (axis cs:5.4,0.910344827586207);
\draw[draw=none,fill=darkorange25512714] (axis cs:5.6,0.773333333333333) rectangle (axis cs:6.4,0.813333333333333);
\draw[draw=none,fill=darkorange25512714] (axis cs:6.6,0.0275862068965517) rectangle (axis cs:7.4,0.324137931034483);
\draw[draw=none,fill=darkorange25512714] (axis cs:7.6,0.754545454545455) rectangle (axis cs:8.4,0.972727272727273);
\draw[draw=none,fill=darkorange25512714] (axis cs:8.6,0.173333333333333) rectangle (axis cs:9.4,0.393333333333333);
\draw[draw=none,fill=darkorange25512714] (axis cs:9.6,0) rectangle (axis cs:10.4,0.26);
\draw[draw=none,fill=redred] (axis cs:0.6,0.966666666666667) rectangle (axis cs:1.4,1);

\draw[draw=none,fill=redred] (axis cs:1.6,0.64) rectangle (axis cs:2.4,1);
\draw[draw=none,fill=redred] (axis cs:2.6,0.938461538461538) rectangle (axis cs:3.4,1);
\draw[draw=none,fill=redred] (axis cs:3.6,1) rectangle (axis cs:4.4,1);
\draw[draw=none,fill=redred] (axis cs:4.6,0.910344827586207) rectangle (axis cs:5.4,1);
\draw[draw=none,fill=redred] (axis cs:5.6,0.813333333333333) rectangle (axis cs:6.4,1);
\draw[draw=none,fill=redred] (axis cs:6.6,0.324137931034483) rectangle (axis cs:7.4,1);
\draw[draw=none,fill=redred] (axis cs:7.6,0.972727272727273) rectangle (axis cs:8.4,1);
\draw[draw=none,fill=redred] (axis cs:8.6,0.393333333333333) rectangle (axis cs:9.4,1);
\draw[draw=none,fill=redred] (axis cs:9.6,0.26) rectangle (axis cs:10.4,1);
\end{axis}

\end{tikzpicture}}
    \vspace{-1.0cm}
    \caption{Results for Problog-FT for different network sizes.}
    \label{fig:problog_ft_by_size}
\end{figure}

\autoref{fig:problog_ft_by_size} and \autoref{fig:gpt4_causalcot_by_size} sort the results of ProbLog-FT and \causalcot on the ten networks in the test in ascending order by number of premises.
There is a clear trend that shows that a growing number of background premises lead to an increasing amount of failure cases.
This can be explained by the fact that having many background premises requires the model to work with an increasingly large message context of all already-parsed ProbLog premises.

\begin{figure}
    \centering
    \scalebox{0.8}{\begin{tikzpicture}

\definecolor{darkgray176}{RGB}{176,176,176}
\definecolor{darkorange25512714}{RGB}{255,127,14}
\definecolor{redred}{RGB}{220, 0, 0}
\definecolor{lightgray204}{RGB}{204,204,204}
\definecolor{steelblue31119180}{RGB}{31,119,180}

\begin{customlegend}[legend columns=5,legend style={align=center,draw=none,column sep=2ex},
width=4.8in,
legend entries={\% Correct, \% Wrong, \% Error}]
\addlegendimage{ybar,ybar legend,draw=none,fill=steelblue31119180}
\addlegendimage{ybar,ybar legend,draw=none,fill=darkorange25512714}
\addlegendimage{ybar,ybar legend,draw=none,fill=redred}
\end{customlegend}
\end{tikzpicture}}

\scalebox{0.8}{\begin{tikzpicture}

\definecolor{darkgray176}{RGB}{176,176,176}
\definecolor{darkorange25512714}{RGB}{255,127,14}
\definecolor{redred}{RGB}{220, 0, 0}
\definecolor{lightgray204}{RGB}{204,204,204}
\definecolor{steelblue31119180}{RGB}{31,119,180}

\begin{axis}[
legend cell align={left},
legend style={fill opacity=0.8, draw opacity=1, text opacity=1, draw=lightgray204},
tick align=outside,
tick pos=left,
x grid style={darkgray176},
xmin=0.11, xmax=10.89,
xtick style={color=black},
y grid style={darkgray176},
ymin=0, ymax=1.06448275862069,
ytick style={color=black},
xtick={1,2,3,4,5,6,7,8,9,10},
xticklabels={2.0, 2.0, 2.0, 2.6, 2.7, 3.0, 3.1, 3.2, 3.3, 6.0},
grid,
xlabel={Avg. Count of States per Variable},
ylabel={Percentage}
]
\draw[draw=none,fill=steelblue31119180] (axis cs:0.6,0) rectangle (axis cs:1.4,0.938461538461538);

\draw[draw=none,fill=steelblue31119180] (axis cs:1.6,0) rectangle (axis cs:2.4,0.76551724137931);
\draw[draw=none,fill=steelblue31119180] (axis cs:2.6,0) rectangle (axis cs:3.4,0.6);
\draw[draw=none,fill=steelblue31119180] (axis cs:3.6,0) rectangle (axis cs:4.4,0.773333333333333);
\draw[draw=none,fill=steelblue31119180] (axis cs:4.6,0) rectangle (axis cs:5.4,0.173333333333333);
\draw[draw=none,fill=steelblue31119180] (axis cs:5.6,0) rectangle (axis cs:6.4,0.65);
\draw[draw=none,fill=steelblue31119180] (axis cs:6.6,0) rectangle (axis cs:7.4,0.754545454545455);
\draw[draw=none,fill=steelblue31119180] (axis cs:7.6,0) rectangle (axis cs:8.4,0);
\draw[draw=none,fill=steelblue31119180] (axis cs:8.6,0) rectangle (axis cs:9.4,0.911111111111111);
\draw[draw=none,fill=steelblue31119180] (axis cs:9.6,0) rectangle (axis cs:10.4,0.0275862068965517);
\draw[draw=none,fill=darkorange25512714] (axis cs:0.6,0.938461538461538) rectangle (axis cs:1.4,0.938461538461538);

\draw[draw=none,fill=darkorange25512714] (axis cs:1.6,0.76551724137931) rectangle (axis cs:2.4,0.910344827586207);
\draw[draw=none,fill=darkorange25512714] (axis cs:2.6,0.6) rectangle (axis cs:3.4,0.64);
\draw[draw=none,fill=darkorange25512714] (axis cs:3.6,0.773333333333333) rectangle (axis cs:4.4,0.813333333333333);
\draw[draw=none,fill=darkorange25512714] (axis cs:4.6,0.173333333333333) rectangle (axis cs:5.4,0.393333333333333);
\draw[draw=none,fill=darkorange25512714] (axis cs:5.6,0.65) rectangle (axis cs:6.4,1);
\draw[draw=none,fill=darkorange25512714] (axis cs:6.6,0.754545454545455) rectangle (axis cs:7.4,0.972727272727273);
\draw[draw=none,fill=darkorange25512714] (axis cs:7.6,0) rectangle (axis cs:8.4,0.26);
\draw[draw=none,fill=darkorange25512714] (axis cs:8.6,0.911111111111111) rectangle (axis cs:9.4,0.966666666666667);
\draw[draw=none,fill=darkorange25512714] (axis cs:9.6,0.0275862068965517) rectangle (axis cs:10.4,0.324137931034483);
\draw[draw=none,fill=redred] (axis cs:0.6,0.938461538461538) rectangle (axis cs:1.4,1);

\draw[draw=none,fill=redred] (axis cs:1.6,0.910344827586207) rectangle (axis cs:2.4,1);
\draw[draw=none,fill=redred] (axis cs:2.6,0.64) rectangle (axis cs:3.4,1);
\draw[draw=none,fill=redred] (axis cs:3.6,0.813333333333333) rectangle (axis cs:4.4,1);
\draw[draw=none,fill=redred] (axis cs:4.6,0.393333333333333) rectangle (axis cs:5.4,1);
\draw[draw=none,fill=redred] (axis cs:5.6,1) rectangle (axis cs:6.4,1);
\draw[draw=none,fill=redred] (axis cs:6.6,0.972727272727273) rectangle (axis cs:7.4,1);
\draw[draw=none,fill=redred] (axis cs:7.6,0.26) rectangle (axis cs:8.4,1);
\draw[draw=none,fill=redred] (axis cs:8.6,0.966666666666667) rectangle (axis cs:9.4,1);
\draw[draw=none,fill=redred] (axis cs:9.6,0.324137931034483) rectangle (axis cs:10.4,1);
\end{axis}

\end{tikzpicture}}
    \vspace{-1.0cm}
    \caption{Results for Problog-FT for various numbers of states.}
    \label{fig:problog_ft_by_states}
\end{figure}
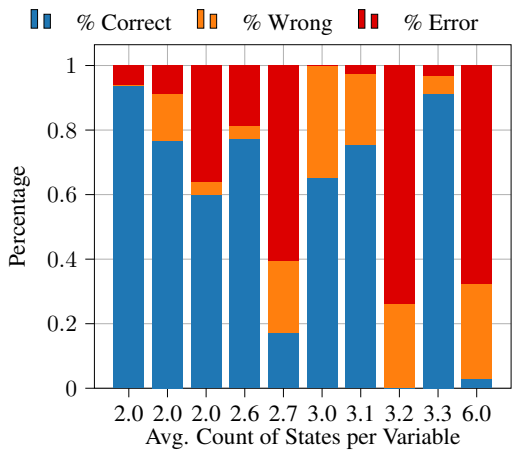

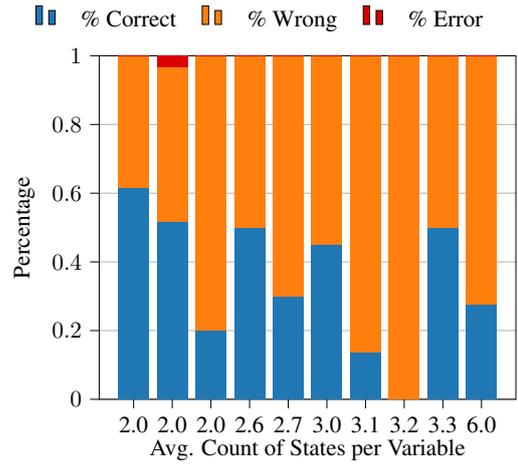
\begin{figure}
    \centering
    \scalebox{0.8}{\begin{tikzpicture}

\definecolor{darkgray176}{RGB}{176,176,176}
\definecolor{darkorange25512714}{RGB}{255,127,14}
\definecolor{redred}{RGB}{220, 0, 0}
\definecolor{lightgray204}{RGB}{204,204,204}
\definecolor{steelblue31119180}{RGB}{31,119,180}

\begin{customlegend}[legend columns=5,legend style={align=center,draw=none,column sep=2ex},
width=4.8in,
legend entries={\% Correct, \% Wrong, \% Error}]
\addlegendimage{ybar,ybar legend,draw=none,fill=steelblue31119180}
\addlegendimage{ybar,ybar legend,draw=none,fill=darkorange25512714}
\addlegendimage{ybar,ybar legend,draw=none,fill=redred}
\end{customlegend}
\end{tikzpicture}}

\scalebox{0.8}{\begin{tikzpicture}

\definecolor{darkgray176}{RGB}{176,176,176}
\definecolor{darkorange25512714}{RGB}{255,127,14}
\definecolor{redred}{RGB}{220, 0, 0}
\definecolor{lightgray204}{RGB}{204,204,204}
\definecolor{steelblue31119180}{RGB}{31,119,180}

\begin{axis}[
legend cell align={left},
legend style={fill opacity=0.8, draw opacity=1, text opacity=1, draw=lightgray204},
tick align=outside,
tick pos=left,
x grid style={darkgray176},
xmin=0.11, xmax=10.89,
xtick style={color=black},
y grid style={darkgray176},
ymin=0, ymax=1,
ytick style={color=black},
xtick={1,2,3,4,5,6,7,8,9,10},
xticklabels={2.0, 2.0, 2.0, 2.6, 2.7, 3.0, 3.1, 3.2, 3.3, 6.0},
grid,
xlabel={Avg. Count of States per Variable},
ylabel={Percentage}
]
\draw[draw=none,fill=steelblue31119180] (axis cs:0.6,0) rectangle (axis cs:1.4,0.615384615384615);
\draw[draw=none,fill=steelblue31119180] (axis cs:1.6,0) rectangle (axis cs:2.4,0.517241379310345);
\draw[draw=none,fill=steelblue31119180] (axis cs:2.6,0) rectangle (axis cs:3.4,0.2);
\draw[draw=none,fill=steelblue31119180] (axis cs:3.6,0) rectangle (axis cs:4.4,0.5);
\draw[draw=none,fill=steelblue31119180] (axis cs:4.6,0) rectangle (axis cs:5.4,0.3);
\draw[draw=none,fill=steelblue31119180] (axis cs:5.6,0) rectangle (axis cs:6.4,0.45);
\draw[draw=none,fill=steelblue31119180] (axis cs:6.6,0) rectangle (axis cs:7.4,0.136363636363636);
\draw[draw=none,fill=steelblue31119180] (axis cs:7.6,0) rectangle (axis cs:8.4,0);
\draw[draw=none,fill=steelblue31119180] (axis cs:8.6,0) rectangle (axis cs:9.4,0.5);
\draw[draw=none,fill=steelblue31119180] (axis cs:9.6,0) rectangle (axis cs:10.4,0.275862068965517);
\draw[draw=none,fill=darkorange25512714] (axis cs:0.6,0.615384615384615) rectangle (axis cs:1.4,1);

\draw[draw=none,fill=darkorange25512714] (axis cs:1.6,0.517241379310345) rectangle (axis cs:2.4,0.96551724137931);
\draw[draw=none,fill=darkorange25512714] (axis cs:2.6,0.2) rectangle (axis cs:3.4,1);
\draw[draw=none,fill=darkorange25512714] (axis cs:3.6,0.5) rectangle (axis cs:4.4,1);
\draw[draw=none,fill=darkorange25512714] (axis cs:4.6,0.3) rectangle (axis cs:5.4,1);
\draw[draw=none,fill=darkorange25512714] (axis cs:5.6,0.45) rectangle (axis cs:6.4,1);
\draw[draw=none,fill=darkorange25512714] (axis cs:6.6,0.136363636363636) rectangle (axis cs:7.4,1);
\draw[draw=none,fill=darkorange25512714] (axis cs:7.6,0) rectangle (axis cs:8.4,1);
\draw[draw=none,fill=darkorange25512714] (axis cs:8.6,0.5) rectangle (axis cs:9.4,1);
\draw[draw=none,fill=darkorange25512714] (axis cs:9.6,0.275862068965517) rectangle (axis cs:10.4,1);
\draw[draw=none,fill=redred] (axis cs:0.6,1) rectangle (axis cs:1.4,1);

\draw[draw=none,fill=redred] (axis cs:1.6,0.96551724137931) rectangle (axis cs:2.4,1);
\draw[draw=none,fill=redred] (axis cs:2.6,1) rectangle (axis cs:3.4,1);
\draw[draw=none,fill=redred] (axis cs:3.6,1) rectangle (axis cs:4.4,1);
\draw[draw=none,fill=redred] (axis cs:4.6,1) rectangle (axis cs:5.4,1);
\draw[draw=none,fill=redred] (axis cs:5.6,1) rectangle (axis cs:6.4,1);
\draw[draw=none,fill=redred] (axis cs:6.6,1) rectangle (axis cs:7.4,1);
\draw[draw=none,fill=redred] (axis cs:7.6,1) rectangle (axis cs:8.4,1);
\draw[draw=none,fill=redred] (axis cs:8.6,1) rectangle (axis cs:9.4,1);
\draw[draw=none,fill=redred] (axis cs:9.6,1) rectangle (axis cs:10.4,1);
\end{axis}

\end{tikzpicture}}
    \vspace{-1.0cm}
    \caption{Results for CausualCoT with GPT4 for various numbers of states.}
    \label{fig:gpt4_causalcot_by_states}
\end{figure}

When analyzing the performance for different numbers of states per (categorical) random variable (cf. \autoref{fig:problog_ft_by_states} and \autoref{fig:gpt4_causalcot_by_states}), one cannot identify a clear trend. An increasing number of states seem more challenging, but we suspect that other factors such as complexity of the domain might play a bigger role.
From this, we conclude that the amount of background premises seems to have a larger influence on the failure probability than the average amount of states per random variable.

\begin{figure}
    \centering
    \scalebox{0.8}{\begin{tikzpicture}

\definecolor{darkgray176}{RGB}{176,176,176}
\definecolor{darkorange25512714}{RGB}{255,127,14}
\definecolor{redred}{RGB}{220, 0, 0}
\definecolor{lightgray204}{RGB}{204,204,204}
\definecolor{steelblue31119180}{RGB}{31,119,180}

\begin{customlegend}[legend columns=5,legend style={align=center,draw=none,column sep=2ex},
width=4.8in,
legend entries={\% Correct, \% Wrong, \% Error}]
\addlegendimage{ybar,ybar legend,draw=none,fill=steelblue31119180}
\addlegendimage{ybar,ybar legend,draw=none,fill=darkorange25512714}
\addlegendimage{ybar,ybar legend,draw=none,fill=redred}
\end{customlegend}
\end{tikzpicture}}

\scalebox{0.8}{\begin{tikzpicture}

\definecolor{darkgray176}{RGB}{176,176,176}
\definecolor{darkorange25512714}{RGB}{255,127,14}
\definecolor{redred}{RGB}{220, 0, 0}
\definecolor{lightgray204}{RGB}{204,204,204}
\definecolor{steelblue31119180}{RGB}{31,119,180}

\begin{axis}[
legend cell align={left},
legend style={fill opacity=0.8, draw opacity=1, text opacity=1, draw=lightgray204},
tick align=outside,
tick pos=left,
x grid style={darkgray176},
xmin=-0.84, xmax=8.84,
xtick style={color=black},
xtick={0,1,2,3,4,5,6,7,8},
xticklabels={9,10,13,16,18,29,37,45,87},
y grid style={darkgray176},
ymin=0, ymax=1,
ytick style={color=black},
grid,
xlabel={Network Size},
ylabel={Percentage}
]
\draw[draw=none,fill=steelblue31119180] (axis cs:-0.4,0) rectangle (axis cs:0.4,0.2);
\draw[draw=none,fill=steelblue31119180] (axis cs:0.6,0) rectangle (axis cs:1.4,0.615384615384615);
\draw[draw=none,fill=steelblue31119180] (axis cs:1.6,0) rectangle (axis cs:2.4,0.45);
\draw[draw=none,fill=steelblue31119180] (axis cs:2.6,0) rectangle (axis cs:3.4,0.517241379310345);
\draw[draw=none,fill=steelblue31119180] (axis cs:3.6,0) rectangle (axis cs:4.4,0.5);
\draw[draw=none,fill=steelblue31119180] (axis cs:4.6,0) rectangle (axis cs:5.4,0.275862068965517);
\draw[draw=none,fill=steelblue31119180] (axis cs:5.6,0) rectangle (axis cs:6.4,0.136363636363636);
\draw[draw=none,fill=steelblue31119180] (axis cs:6.6,0) rectangle (axis cs:7.4,0.3);
\draw[draw=none,fill=steelblue31119180] (axis cs:7.6,0) rectangle (axis cs:8.4,0);
\draw[draw=none,fill=darkorange25512714] (axis cs:-0.4,0.5) rectangle (axis cs:0.4,1);

\draw[draw=none,fill=darkorange25512714] (axis cs:-0.4,0.2) rectangle (axis cs:0.4,1);
\draw[draw=none,fill=darkorange25512714] (axis cs:0.6,0.615384615384615) rectangle (axis cs:1.4,1);
\draw[draw=none,fill=darkorange25512714] (axis cs:1.6,0.45) rectangle (axis cs:2.4,1);
\draw[draw=none,fill=darkorange25512714] (axis cs:2.6,0.517241379310345) rectangle (axis cs:3.4,0.96551724137931);
\draw[draw=none,fill=darkorange25512714] (axis cs:3.6,0.5) rectangle (axis cs:4.4,1);
\draw[draw=none,fill=darkorange25512714] (axis cs:4.6,0.275862068965517) rectangle (axis cs:5.4,1);
\draw[draw=none,fill=darkorange25512714] (axis cs:5.6,0.136363636363636) rectangle (axis cs:6.4,1);
\draw[draw=none,fill=darkorange25512714] (axis cs:6.6,0.3) rectangle (axis cs:7.4,1);
\draw[draw=none,fill=darkorange25512714] (axis cs:7.6,0) rectangle (axis cs:8.4,1);
\draw[draw=none,fill=redred] (axis cs:-0.4,1) rectangle (axis cs:0.4,1);

\draw[draw=none,fill=redred] (axis cs:-0.4,1) rectangle (axis cs:0.4,1);
\draw[draw=none,fill=redred] (axis cs:0.6,1) rectangle (axis cs:1.4,1);
\draw[draw=none,fill=redred] (axis cs:1.6,1) rectangle (axis cs:2.4,1);
\draw[draw=none,fill=redred] (axis cs:2.6,0.96551724137931) rectangle (axis cs:3.4,1);
\draw[draw=none,fill=redred] (axis cs:3.6,1) rectangle (axis cs:4.4,1);
\draw[draw=none,fill=redred] (axis cs:4.6,1) rectangle (axis cs:5.4,1);
\draw[draw=none,fill=redred] (axis cs:5.6,1) rectangle (axis cs:6.4,1);
\draw[draw=none,fill=redred] (axis cs:6.6,1) rectangle (axis cs:7.4,1);
\draw[draw=none,fill=redred] (axis cs:7.6,1) rectangle (axis cs:8.4,1);
\end{axis}

\end{tikzpicture}}
    \vspace{-1.0cm}
    \caption{Results for CausualCoT with GPT4 for different network sizes.}
    \label{fig:gpt4_causalcot_by_size}
\end{figure}
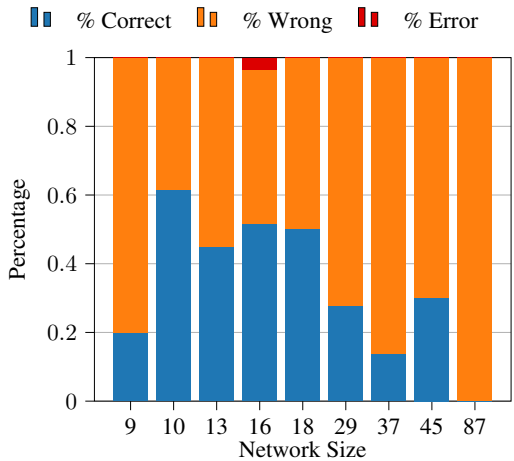

\begin{figure*}
    \centering
    \includegraphics[width=1.0\linewidth]{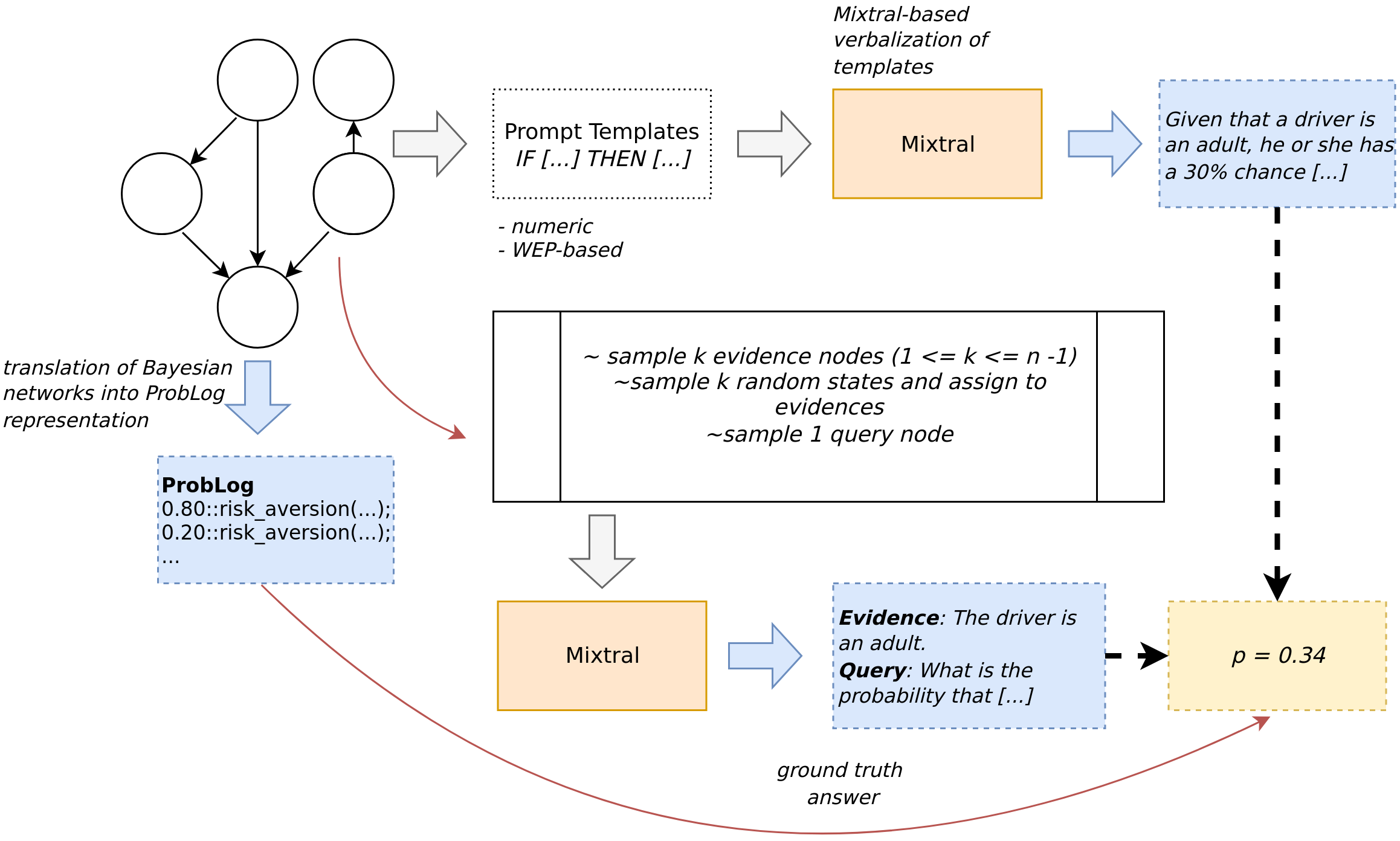}
    \caption{An schematic overview of our data generation pipeline.}
    \label{fig:data_generation}
\end{figure*}

\section{Data Generation Pipeline}
\label{sec:data_generation_pipeline}
\autoref{fig:data_generation} displays an overview of our data generation pipeline.
Every CPT entry is verbalized using the Mixtral LLM.
Randomly sampled evidences and question nodes, which build the QE pairs in \datasetname, are also transferred into natural language using Mixtral.
For that, we prompt the LLM with the instruction to formulate grounded statements in the case of evidences and questions asking for probabilities in the case of queries.
For each BN, we create an equivalent ProbLog representation that is used to generate the ground truth answer and to fine-tune the LLMs.

\section{\cladder and \blind}
\label{sec:related_datasets}
In this section, we provide one sample from \cladder and \blind each.

\cladder uses pre-defined BN structures with three or four nodes.
The following background premises are taken from a three-node network and represent the distribution over two binary random variables that can take the values \textit{true} or \textit{false}:
\begin{quote}
\textit{The overall probability of alarm set by husband is 3\%. For husbands that don't set the alarm, the probability of ringing alarm is 74\%. For husbands that set the alarm, the probability of ringing alarm is 22\%.}  
\end{quote}
The corresponding question \textit{Is ringing alarm more likely than silent alarm overall?} requires the following computation to obtain the answer: $\mathbb{P}(alarm = ringing) > \mathbb{P}(alarm = silent)$

One instance in \blind is the following set of background premises:
\begin{quote}
\textit{If purple event is False, then grey event is True with probability of 39\%. If purple event is False, then grey event is False with probability of 61\%. If purple event is True, then grey event is True with probability of 3\%. If purple event is True, then grey event is False with probability of 97\%. purple event is true with probability of 55\%. purple event is false with probability of 45\%.}
\end{quote}
The corresponding question \textit{What is the probability that grey event is True given that purple event is False?} requires the following computation to obtain the answer: $\mathbb{P}(grey = True | purple = False)$.
Again, all random variables, in this case the color events, can only take the two states \textit{true} or \textit{false}.

\end{document}